\documentclass[letterpaper, 10 pt, conference]{ieeeconf}
\IEEEoverridecommandlockouts
\usepackage{amsmath}
\usepackage{amssymb}
\usepackage{enumerate}
\usepackage{algorithm}
\usepackage{algpseudocode}
\usepackage{graphicx}
\usepackage{hyperref}
\usepackage{cite}
\usepackage{sidecap}
\usepackage{booktabs}
\usepackage{algorithm}
\usepackage{algorithm}
\usepackage{algpseudocode}

\usepackage{balance}

\graphicspath{./images}
\DeclareGraphicsExtensions{.pdf,.eps, .jpg, .png, .PNG}
\title{\LARGE \bf MPBRS: Multi-Potential-Based Reward Shaping for Unified Reward Configuration in Humanoid Locomotion}
\author{
Van Chung Nguyen$^1$, and Hung Manh La $^1$   
\thanks{$^1$are with the Advanced Robotics and Automation (ARA) Lab, Department of Computer Science and Engineering, University of Nevada, Reno, NV 89557, USA.}
\thanks{Corresponding author: Hung La, email: {\tt\small hla@unr.edu}}
\thanks{Source codes and videos are available at: \url{https://github.com/aralab-unr/MPBRS}}
}
\begin{document}
\maketitle
\setlength{\textfloatsep}{0pt}
\begin{abstract} 
Designing and tuning reward functions remains one of the main challenges in building effective reinforcement learning (RL) pipelines for humanoid locomotion. While shaping rewards can accelerate learning, naively formulated terms often conflict with the desired behavior, overfit to a specific robot, or become unstable under small weight changes. Potential-Based Reward Shaping (PBRS) offers an alternative solution: shaping the reward via differences of a potential function to guide exploration without changing the optimal policy. In this paper, we study Multi-PBRS for humanoid locomotion and propose a unified reward configuration. We then evaluate the same reward structure and weights across nine humanoid robots, requiring only robot-specific metadata (joint/link mappings and default poses) rather than per-robot reward redesign. Our experiments show that, compared to standard shaping, MPBRS provides more consistent scaling behavior and reduced tuning sensitivity across robots, making it easier to tune without altering the optimal policy.
\end{abstract}
\section{INTRODUCTION}
Recently, reinforcement learning (RL) has become one of the most widely used approaches for humanoid locomotion. The use of RL for humanoid locomotion is due to its ability to learn coordinated, whole-body behaviors that are difficult to design by hand \cite{ng1999policy,xie2020learning}. However, despite the success of RL in humanoid locomotion, its application depends heavily on the reward function design \cite{ng1999policy,xie2020learning}. The design of the reward function greatly affects the behavior of the humanoid robot. For example, when the reward is specified in a naive way-such as a quadratic tracking error or a sparse task-success signal-the agent may learn strategies that maximize the reward while producing unintended behaviors that do not match the designer's intent \cite{muller2017quadratic, jeon2023benchmarking}. Conversely, well-shaped rewards can substantially speed up learning, whereas poorly scaled or conflicting terms can cause brittle behaviors, reward hacking, unstable training, or policies that only work for a single morphology and fail under small changes in dynamics or sensing. As a result, developers often spend significant time tuning reward terms and weights to reach an acceptable balance between training stability and behavioral quality. This tuning process is further complicated by the sensitivity of humanoid RL to hyperparameters and reward scaling, which makes it difficult to isolate the effect of individual reward terms and tune them systematically. Furthermore, the design of the reward function is severe for humanoid locomotion since it involves high-dimensional control and contact-rich dynamics with foot-ground interactions, and morphology-dependent constraints. Consequently, many locomotion pipelines rely on extensive reward design and per-robot tuning: weights are repeatedly adjusted until training stabilizes and the resulting gait appears plausible. Unfortunately, this workflow is time-consuming, hard to reproduce, and often does not transfer even between similar humanoids. A reward that works well on one robot may require significant modification on another due to differences in morphology, actuation, or physical design.
\begin{figure} 
    \centering
    \includegraphics[width=1.00\linewidth]{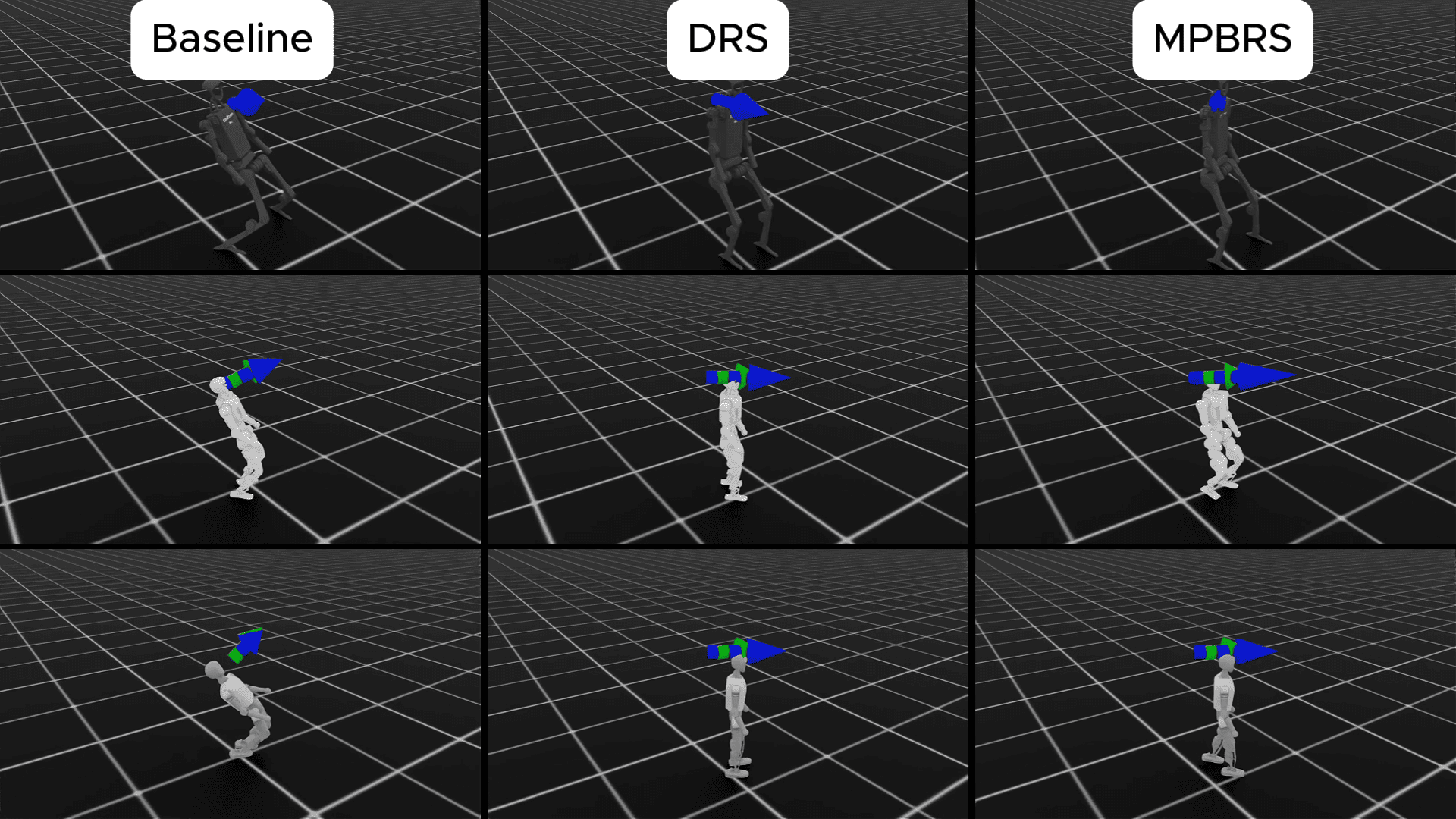}
    \caption{The baseline (left), DRS (middle), and MPBRS (right) locomotion policies are shown. MPBRS succeeds in achieving an optimal policy similar to the carefully tuned DRS, while the baseline leans the base backward during locomotion.}
    \label{compareloco}
\end{figure}
In \cite{ng1999policy}, Ng, Harada, and Russell proposed a class of shaping rewards defined by differences of a potential function that is policy-invariant in discounted MDPs. This method has appeared in many researchers because it can provide richer learning signals while leaving the baseline task objective unchanged, effectively separating task specification from convergence acceleration.

Later, several follow-up works \cite{wiewiora2003potential,devlin2011theoretical,grzes2017reward} examined PBRS in RL and concluded that it leads to faster learning. However, this evidence is largely drawn from grid-worlds or other low-dimensional domains. Meanwhile, high-performing RL results in robotics-especially for legged locomotion-still depend on straightforward, manually tuned shaping rewards that provide no policy-invariance guarantees \cite{radosavovic2024real}.

In this paper, we examine deep RL locomotion in 3D under a unified reward design and test how multi-PBRS compares to standard shaping when deployed unchanged on nine humanoid robots. Rather than delivering large improvements in convergence rate, MPBRS primarily improves training stability and portability, showing lower sensitivity to weight selection and reward magnitude. This reduces tuning overhead and supports more consistent reuse across diverse humanoid designs.

The rest of the paper is organized as follows. Chapter II covers the literature review, outlining the related work and providing an overview of the method. Chapter III presents the methodology of the potential-based reward shaping used in this paper. Chapter IV provides a system overview, including the case study of nine humanoid robots and the experimental setup used in this paper. Chapter V presents the results of the method. Finally, Chapter VI concludes the paper.
\begin{figure*} [!t]
    \centering
    \includegraphics[width=1.0\linewidth]{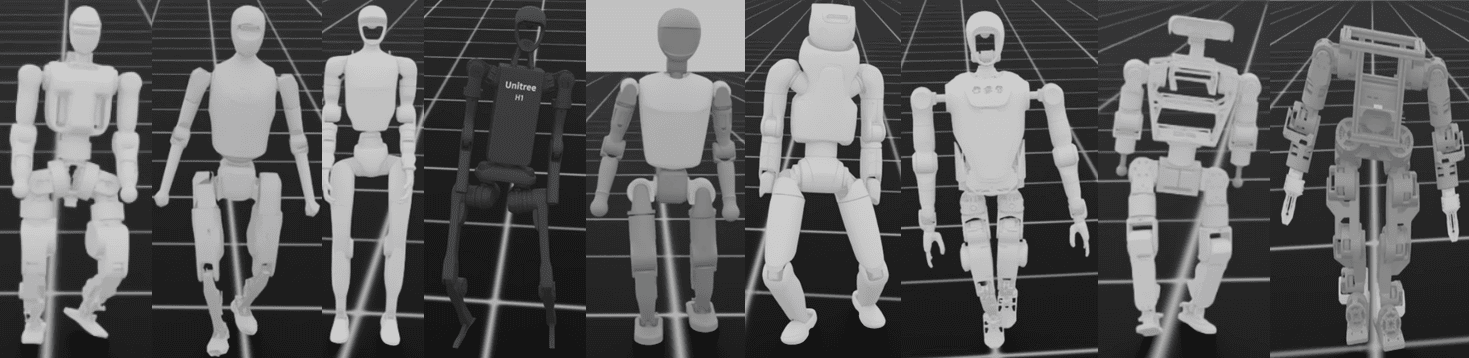}
    \caption{A snapshot of nine humanoid robots used in this paper in IsaacLab. From right to left: Fourier N1, Tienkung2, Unitree G1, Unitree H1, Booster T1, Fourier GR3V2, Kuavo S45, Hightorque HI, and Berkeley Humanoid Lite.}
    \label{humanoidtype}
\end{figure*}
\section{LITERATURE REVIEW}
\subsection{Reward shaping for locomotion}
Recently, high-performing RL locomotion systems often rely on direct reward shaping (DRS). The DRS is a dense terms that encode task-relevant objectives such as velocity tracking, uprightness, base-height regulation, and regularization penalties on energy or action changes \cite{hwangbo2019learning,rudin2022learning,lee2020learning,yang2022fast}. The DRS is attractive because they are simple to compute and do not require external demonstrations, but their effectiveness depends strongly on careful scaling and balancing of terms. In humanoid locomotion, small changes in reward weights can produce different behaviors, ranging from overly conservative gaits to unstable oscillations. The small changes in reward weight can lead to solutions that exploit a proxy objective rather than the intended task \cite{amodei2016concrete,ratner2018simplifying,srisuchinnawong2025gain,song2021deep}. 

A second common approach is imitation learning, which provides dense rewards by tracking \emph{reference signals}. Imitation learning strongly encourages the robot to replicate the desired behavior. In physics-based character control, imitation and reference tracking have been particularly successful for producing coordinated whole-body motion and locomotion skills \cite{peng2018deepmimic, peng2021amp, chentanez2018physics, peng2017deeploco}. In robotics-oriented biped learning, iterative design pipelines often incorporate structured objectives and reference-like heuristics (e.g., posture targets, height/orientation regulation, cadence-related signals) and emphasize that reward shaping and tuning are major determinants of training stability and final gait quality \cite{xie2018feedback,li2021reinforcement}. While imitation learning can substantially accelerate learning, it introduces an additional design burden: the choice of reference quantities and their weights can bias solutions toward reproducing the reference rather than solving the underlying task, and these choices frequently do not transfer cleanly across morphologies \cite{peng2020learning,tang2024humanmimic,mirza2025imitation}. This lack of portability motivates shaping methods that provide dense guidance while reducing sensitivity to hand-tuned scaling.

\subsection{Potential-based reward shaping (PBRS)}
Ng, Harada, and Russell \cite{ng1999policy} proposed \emph{potential-based reward shaping} (PBRS), showing that shaping rewards constructed in this way preserve the set of optimal policies in discounted MDPs. This property has appealed to many researchers, as it suggests a principled separation between (i) a base reward that expresses the task objective and (ii) an auxiliary shaping signal that improves the robot's behaviour without altering an optimal solution. Wiewiora \cite{wiewiora2003potential} further showed that PBRS is closely related to $Q$-value initialization, providing intuition for why PBRS can speed up early learning: an informative potential effectively redistributes long-horizon value into denser step-wise feedback. Subsequent work has explored PBRS in broader settings (e.g., episodic tasks and learned or dynamic potentials) and generally reports faster convergence in low-dimensional environments such as gridworlds or classic control \cite{devlin2012dynamic,harutyunyan2015expressing}. 

For humanoid locomotion, however, the primary practical benefit of PBRS appears to differ from the common expectation of large convergence gains. Jeon \emph{et al.} \cite{jeon2023benchmarking} benchmark PBRS-style shaping in humanoid locomotion and find that improvements in convergence speed can be modest, while PBRS terms are notably more robust to reward scaling than typical direct shaping. This motivates treating PBRS not only as a tool for faster learning, but as a mechanism for \emph{reducing tuning sensitivity} and improving reward portability. Building on this perspective, our work evaluates a \emph{single unified multi-PBRS reward configuration} reused across nine humanoid robots, with the goal of measuring how MPBRS affects robustness to scaling and the degree of per-robot reward retuning required.

\section{METHODOLOGY}

\subsection{Markov Decision Processes and Reinforcement Learning}
\label{sec:mdp_rl}
We formalize humanoid locomotion as a Markov decision process (MDP): $\mathcal{M}=(\mathcal{S}, \mathcal{A}, \mathcal{T}, r, \gamma)$, where $\mathcal{S}$ is the state space, $\mathcal{A}$ is the action space, $\mathcal{T}(s_{t+1}\mid s_t,a_t)$ defines the system dynamics, $r$ is the reward function, and $\gamma\in(0,1]$ is the discount factor. At each timestep, the agent selects an action $a_t=\pi(s_t)$ according to a policy $\pi$, transitions to $s_{t+1}$, and receives a scalar reward $r(s_t,a_t,s_{t+1})\in\mathbb{R}$. The goal of reinforcement learning is to find a policy that maximizes the expected discounted return:
\begin{equation}
J(\pi) \;=\; \mathbb{E}_{\pi}\left[\sum_{t=0}^{\infty}\gamma^{t}\, r(s_t,a_t,s_{t+1})\right].
\end{equation}
Equivalently, for a fixed policy $\pi$, we define the value function
\begin{equation}
V^{\pi}(s) \;=\; \mathbb{E}_{\pi}\left[\sum_{t=0}^{\infty}\gamma^{t}\, r(s_t,a_t,s_{t+1}) \,\middle|\, s_0=s\right],
\end{equation}
and seek $\pi^\ast$ that maximizes $V^{\pi}(s)$ for all $s$.

\subsection{Reward shaping: DRS vs.\ Multi-PBRS}
\label{sec:reward_shaping}

Reward shaping is widely used to make reinforcement learning for locomotion practical by providing dense feedback beyond a sparse or weak task objective. Let $r(s_t,a_t,s_{t+1})$ be the baseline reward for the transition $(s_t,a_t,s_{t+1})$. We add a shaping term $R(s_t,a_t,s_{t+1})$ and obtain the shaped reward
\begin{equation}
\hat r(s_t,a_t,s_{t+1}) \;=\; r(s_t,a_t,s_{t+1}) \;+\; R(s_t,a_t,s_{t+1}),
\label{eq:shaped_reward}
\end{equation}
which induces a shaped MDP $\mathcal{M}_{\text{shaped}}=(\mathcal{S},\mathcal{A},\mathcal{T},\hat r,\gamma)$. Dense shaping can substantially improve convergence, but it also introduces the risk that the learned policy optimizes the shaping proxy rather than the intended task, producing unintended behaviors when the shaping terms are mis-specified or mis-scaled~\cite{jeon2023benchmarking,amodei2016concrete}.
\begin{table}
\centering
\caption{Humanoid Robots Overview.}
\label{tab:humanoidrobot}
\begin{tabular}{lccc}
\hline
\textbf{Name} & \textbf{Joints Dims} & \textbf{Actions Dims} & \textbf{Waist Dims}\\
\hline
\text{Fourier N1}              & 23 & 13 & 1 \\
\text{Unitree H1}              & 19 & 11 & 1 \\
\text{Unitree G1}              & 29 & 13 & 3 \\
\text{Booster T1}              & 23 & 13 & 1 \\
\text{Fourier GR3V2}           & 31 & 13 & 3 \\
\text{HighTorque HI}           & 25 & 13 & 3 \\
\text{Kuavo S45}               & 28 & 12 & None \\
\text{TienKung2}               & 20 & 12 & None \\
\text{Berkeley Lite}  & 22 & 12 & None \\
\hline
\end{tabular}
\end{table}
\paragraph{Direct reward shaping (DRS)}
The most common shaping strategy in robotics is to add instantaneous terms that are easy to compute from the current state (and sometimes the action). Let $\Phi_i:\mathcal{S}\rightarrow\mathbb{R}$ denote scalar signals designed to correlate with desirable locomotion properties such as velocity tracking, uprightness, nominal posture, smooth actions, or foot stability. Direct shaping adds these signals at each timestep:
\begin{align}
&R_{\text{DRS}}(s_t,a_t,s_{t+1}) \;=\; \sum_{i=1}^{K} w_i\,\Phi_i(s_t) \notag \\
&\hat r_{\text{DRS}} \;=\; r \;+\; R_{\text{DRS}} .
\label{eq:drs}
\end{align}
DRS has attracted a lot of research interest due to its simplicity in deployment, provides immediate feedback, and can be implemented using only onboard quantities (proprioception, IMU, contacts). However, DRS is often \emph{weight-sensitive}: the magnitude of each $\Phi_i$ depends on the robot morphology, controller gains, contact regime, and sensor conventions. These magnitudes can vary across training as the policy changes. As a result, small changes in $w_i$ can cause huge different behaviors (e.g., overly cautious gaits, oscillations, collapsing into `standing still' solutions, or exploiting a proxy signal), especially in contact-rich humanoid domains where multiple objectives compete (stability, speed, energy, and foot clearance). This sensitivity turns reward design into an iterative tuning process and often prevents straightforward transfer of a reward configuration across different humanoids~\cite{jeon2023benchmarking}.
\paragraph{Potential-based reward shaping (PBRS)}
PBRS uses the same potentials but provides shaping reward based on \emph{progress} in each potential across transitions:
\begin{align}
& R_{\text{PBRS}}(s_t,a_t,s_{t+1})
\;=\;
\sum_{i=1}^{K} w_i\left(\gamma_{\Phi}\,\Phi_i(s_{t+1}) - \Phi_i(s_t)\right) \notag \\
& \hat r_{\text{PBRS}} \;=\; r \;+\; R_{\text{PBRS}},
\label{eq:pbrs}    
\end{align}
where $\gamma_{\Phi}\in(0,1]$ is a shaping discount factor, $\omega_i$ is the weight of the reward functions. Compared to \eqref{eq:drs}, PBRS shifts the learning signal from `being in a good state' to `moving toward a better state.' In particular, the shaping contribution is near zero when the potential is stationary and becomes positive only when the next state increases the potential. This transition-based structure tends to produce shaping signals that are less dominated by absolute offsets and more directly tied to action consequences, which can make learning less brittle when potentials differ in baseline magnitude across morphologies.

From this perspective, PBRS is appealing to researchers who seek to preserve optimal policies while improving learning through the reshaping of intermediate rewards~\cite{ng1999policy}. In practice, however, PBRS does not automatically guarantee faster convergence in high-dimensional locomotion: when the potentials are imperfect or only weakly predictive of long-horizon success, PBRS may provide only modest acceleration. Empirical benchmarking in humanoid locomotion suggests that a more consistent practical advantage of PBRS is improved robustness to scaling and weight selection, which reduces the tuning burden even when convergence speed gains are marginal~\cite{jeon2023benchmarking}.
\begin{table}[t]
\centering
\caption{Agent Observations.}
\label{tab:obs_noise}
\begin{tabular}{lcc}
\hline
\textbf{Observation term} & \textbf{Dim.} & \textbf{Noise / corruption} \\
\hline
Commands $(v_x, v_y, \omega_z)$ & 3 & $0.02$ \\
Base linear velocity (body) $\mathbf{v}_b$ [m/s] & 3 & $0.04$ \\
Base angular velocity (body) $\boldsymbol{\omega}_b$ [rad/s] & 3 & $0.025$ \\
Projected gravity (body) $\hat{\mathbf{g}}_b$ & 3 & $0.015$  \\
Joint position offsets $\mathbf{q}-\mathbf{q}_0$ [rad] & $N_a$ & $0.004$ \\
Joint velocities $\dot{\mathbf{q}}$ [rad/s] & $N_a$ & $0.02$ \\
Previous action $a_{t-1}$ & $N_a$ & $0.015$ \\
\hline
Feet contact (left,right) $b_c$ & 2 & bit-flip prob $= 0.01$ \\
Left foot speed (world xy) [m/s]  & 2 & $0.04$ \\
Right foot speed (world xy) [m/s]  & 2 & $0.04$ \\
\hline
\end{tabular}
{\footnotesize \textbf{Note:} The critic network additionally receives the extra terms in the lower block (feet contact and foot speeds), while the policy network uses only the upper block.}
\end{table}
\paragraph{Multi-PBRS}
Humanoid locomotion typically requires shaping multiple aspects of behavior simultaneously (tracking, posture, upright, and symmetry). For this reason, we adopt \emph{multi-PBRS}, in which several complementary potentials are shaped in parallel using \eqref{eq:pbrs}. Concretely, our unified configuration uses three potentials shared across all robots:
\textbf{(i)} a base-height potential computed from the ray-based ground distance, 
\textbf{(ii)} an uprightness potential computed from the gravity vector.
\textbf{(iii)} a joint-regularization potential that encourages left--right symmetry and nominal hip yaw through robot-specific joint-name mappings.
Each potential is computed online and cached across timesteps so that $\Phi_i(s_t)$ and $\Phi_i(s_{t+1})$ are paired consistently; shaping is set to zero on reset steps to avoid spurious impulses at episode boundaries.
This \emph{multi-PBRS} construction is central to our goal: enabling a single reward design to remain usable across diverse humanoid morphologies with minimal per-robot modification (limited to joint-name dictionaries and nominal height targets).

\section{SYSTEM OVERVIEW}
To examine whether MPBRS remains advantageous in continuous-control, contact-rich, and high-dimensional locomotion, we evaluate it within a single unified training pipeline spanning nine simulated humanoid robots with substantially different morphologies. Rather than redesigning rewards per robot, we hold the learning setup fixed and isolate the effect of shaping by directly comparing the DRS against the MPBRS terms under the same task objectives. This experimental design prioritizes cross-morphology generality: any observed performance differences can be attributed to the shaping formulation rather than to robot-specific reward engineering.
\subsection{Case study}
We evaluate our method across nine humanoid robots, as summarized in the table~\ref{tab:humanoidrobot}. For learning running locomotion, we fix the arm joints at nominal angles and reduce control of the system to only the legs and the waist yaw. If the waist has roll, pitch, and yaw joints, we set the waist roll and pitch to their default values and control only the yaw. 
\subsection{Low-level control (PD tracking)}
The policy outputs normalized actions $\mathbf{a}\in\mathbb{R}^{n_a}$ for the action joints.
We map these to desired joint position targets. These targets are tracked by a joint-space PD controller that computes torques
\begin{align}
\tau_j
= K_{p,j}\bigl(q^{\mathrm{des}}_j-q_j\bigr)
- K_{d,j}\dot q_j,
\label{eq:pd_torque}
\end{align}
where $q_j,\dot q_j$ are the current joint angle and velocity.
Instead of using constant gains for all joints, we tune per-joint gains according to actuator characteristics across joints spanning from the hips to the feet. This tuning ensures smooth robot motion while keeping the control inputs within actuator limits.
\begin{table}[t]
\centering
\caption{Training Rewards.}
\label{tab:training_rewards}
\renewcommand{\arraystretch}{1.15}
\begin{tabular}{p{2.75cm} c p{3.75cm}}
\hline
\textbf{Baseline} & \textbf{Weight} & \textbf{Function} \\
\hline
Linear velocity tracking & $0.5$ &
$\exp\!\left(-\|{\bf v}^{cmd}_{xy}-{\bf v}^{yaw}_{xy}\|^2/\sigma_v^2\right)$ \\
Yaw rate tracking & $0.3$ & $\exp\!\left(-(\omega^{cmd}_{z}-\omega_{z})^2/\sigma_\omega^2\right)$ \\
Waist default & $0.25$ & $\exp\!\left(-8\,|q_{waist}-q_{waist,0}|\right)$  \\
Action magnitude penalty & $-0.005$ &$-\sum_i a_i^2$ \\
Action rate penalty & $-0.05$ & $-\sum_i (a_i-a_{i,prev})^2$ \\
Termination & $-200.0$ & $ \begin{cases}
1, & h < h_{\min},\\
1, & \theta_{\text{tilt}} > \theta_{\max},\\
1, & \text{self-collision},\\
1, & \text{time-out},\\
0, & \text{otherwise},
\end{cases}$ \\
\hline
\textbf{Direct Shaping} & \textbf{Weight} & \textbf{Function} \\
\hline
Height $\Phi_h(s)$ & $0.75$ & $\exp\!\left(-\frac{(h(s)-z_{des})^2}{\sigma_h^2}\right)$ \\
Uprightness $\Phi_{up}(s)$ & $0.5$ &
$\exp\!\left(-4\,|g_b^z + 1|\right)$ \\
Joint regularization $\Phi_{jr}$ & $0.25$ &
$\exp{(-\frac{(q_{La}-q_{Ra})^2}{\sigma_q})}+\exp{(-\frac{(q_{Lp}-q_{Rp})^2}{\sigma_q})}+\exp{(-\frac{q_{Ly}^2}{\sigma_q})}+\exp{(-\frac{q_{Ry}^2}{\sigma_q})}$ \\
\hline
\textbf{Multi-PBRS} & \textbf{Weight} & \textbf{Function} \\
\hline
Height PBRS $R_h$ & $1.0$ &
$R_h=\gamma_{\Phi}\Phi_h(s_{t+1})-\Phi_h(s_t)$ \\
Uprightness PBRS $R_{up}$ & $1.0$ &
$R_{air}=\gamma_{\Phi}\Phi_{up}(s_{t+1})-\Phi_{up}(s_t)$ \\
Joint regularization PBRS $R_{jr}$ & $1.0$ &
$R_{jr}=\gamma_{\Phi}\Phi_{jr}(s_{t+1})-\Phi_{jr}(s_t)$ \\
\hline
\end{tabular}
\end{table}
\begin{table}[t]
\centering
\caption{Training Environment Hyperparameters.}
\label{tab:train_hyperparams}
\begin{tabular}{l c}
\hline
\textbf{Hyperparameter} & \textbf{Value} \\
\hline
Initial action noise std & 1.0 (log) \\
Value loss coefficient & 1.0 \\
Clipping $\epsilon$ & 0.2 \\
Entropy coefficient & 0.001 \\
Learning rate & $3\times10^{-4}$ (adaptive) \\
Discount factor $\gamma$ & 0.99 \\
$\lambda$ (GAE) & 0.95 \\
Steps/env & 64 \\
Learning epochs & 8 \\
Mini-batches & 16 \\
Desired KL & 0.01 \\
Max grad norm & 1.0 \\
Policy network size & [512, 256, 128, 64] \\
Critic network size & [512, 256, 128, 64] \\
Activation & ELU \\
\hline
\end{tabular}
\end{table}
\subsection{Training setup}
For the training setup, we define the observation space as shown in Table~\ref{tab:obs_noise}. Its dimension is $18 + 3N_a$, where $N_a$ depends on the humanoid robot type. We design the observations to minimize reliance on external sensors, which simplifies future real-world deployment. To model sensor imperfections during training, we corrupt the observations with uniformly sampled noise (Table~\ref{tab:obs_noise}). This helps the policy become more robust to measurement errors.
For the humanoid reward design, the baseline locomotion reward is defined in Table~\ref{tab:training_rewards}. In this baseline, only the first two terms-linear velocity tracking and yaw-rate tracking-are directly task-related. The remaining terms act as regularizers to promote smoother motion and encourage energy-efficient locomotion.
The tracking parameters are set to $\sigma_v = \sigma_{\omega} = 0.5$. Termination thresholds are chosen based on the robot type and the nominal height of its \texttt{base\_link}.
For direct reward shaping (DRS), we define three terms analogous to those used in MPBRS, and the corresponding DRS functions are also listed in Table~\ref{tab:training_rewards}. The reward tuning was performed to obtain moderate robot behavior.
For the potential-based reward shaping (PBRS), each shaping objective we define a scalar potential function $\phi(s)$ and convert it into a shaped reward as in \eqref{eq:pbrs}. In our experiments, we set $\gamma_\phi = 1.00$ for all PBRS terms to improve training stability and speed, as discussed in \cite{grzes2009theoretical, malysheva2018learning, devlin2011theoretical, jeon2023benchmarking}. Our findings are consistent with prior work and further support the use of an undiscounted potential in PBRS. As discussed before, we choose three shaping rewards commonly used in humanoid locomotion:
\begin{enumerate}
\item \textbf{Height potential} $\phi_h(s)$: We set $\sigma_h = 0.08$.
\item \textbf{Uprightness potential} $\phi_{\text{up}}(s)$: We use the gravity vector component $g_b^z$ to encourage locomotion when the robot base remains upright.
\item \textbf{Joint-regularization potential} $\phi_j(s)$: We set $\sigma_q = 0.5$.
\end{enumerate}
\begin{figure*} 
    \centering
    \includegraphics[width=0.315\linewidth]{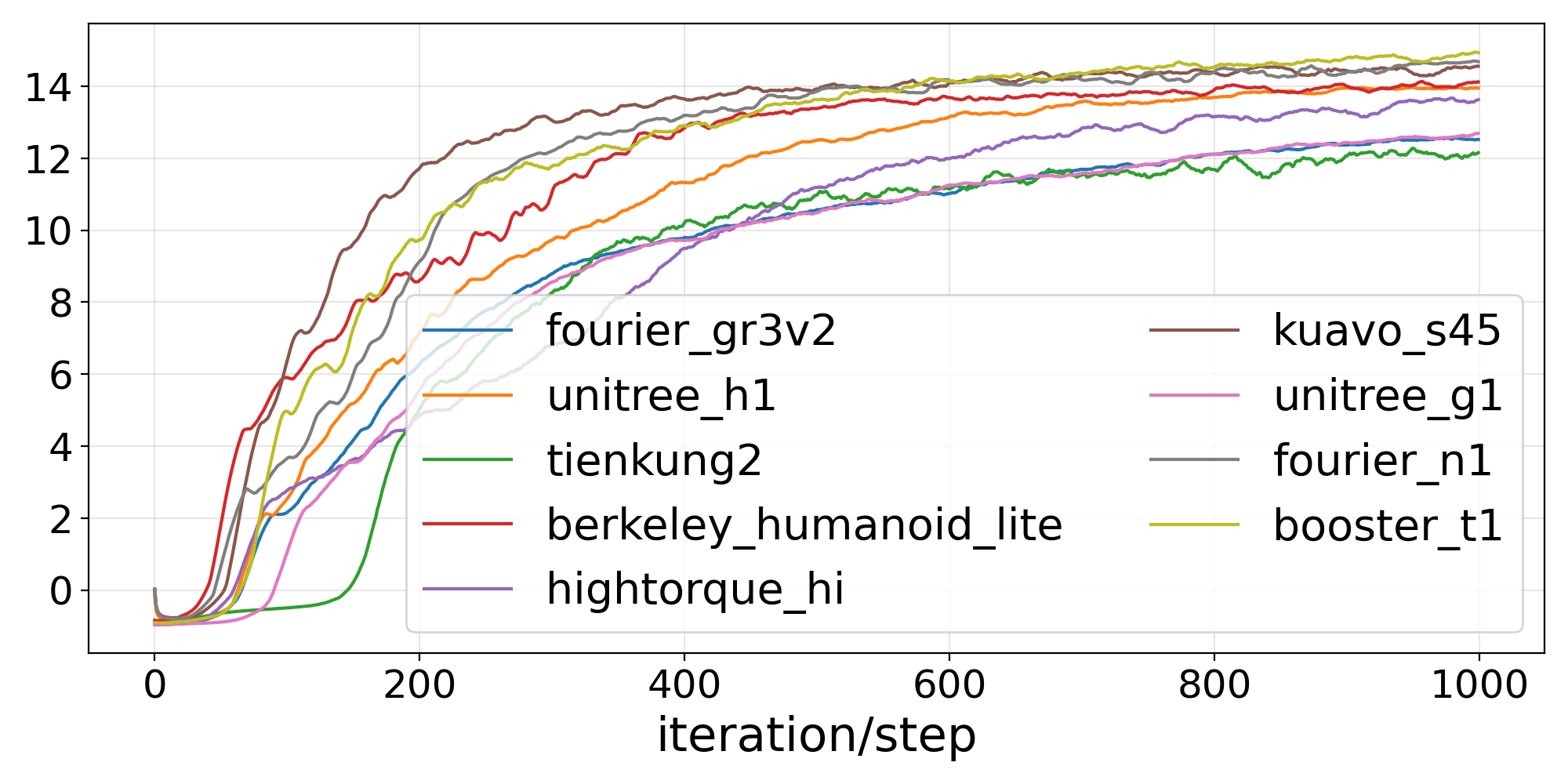}
    \includegraphics[width=0.315\linewidth]{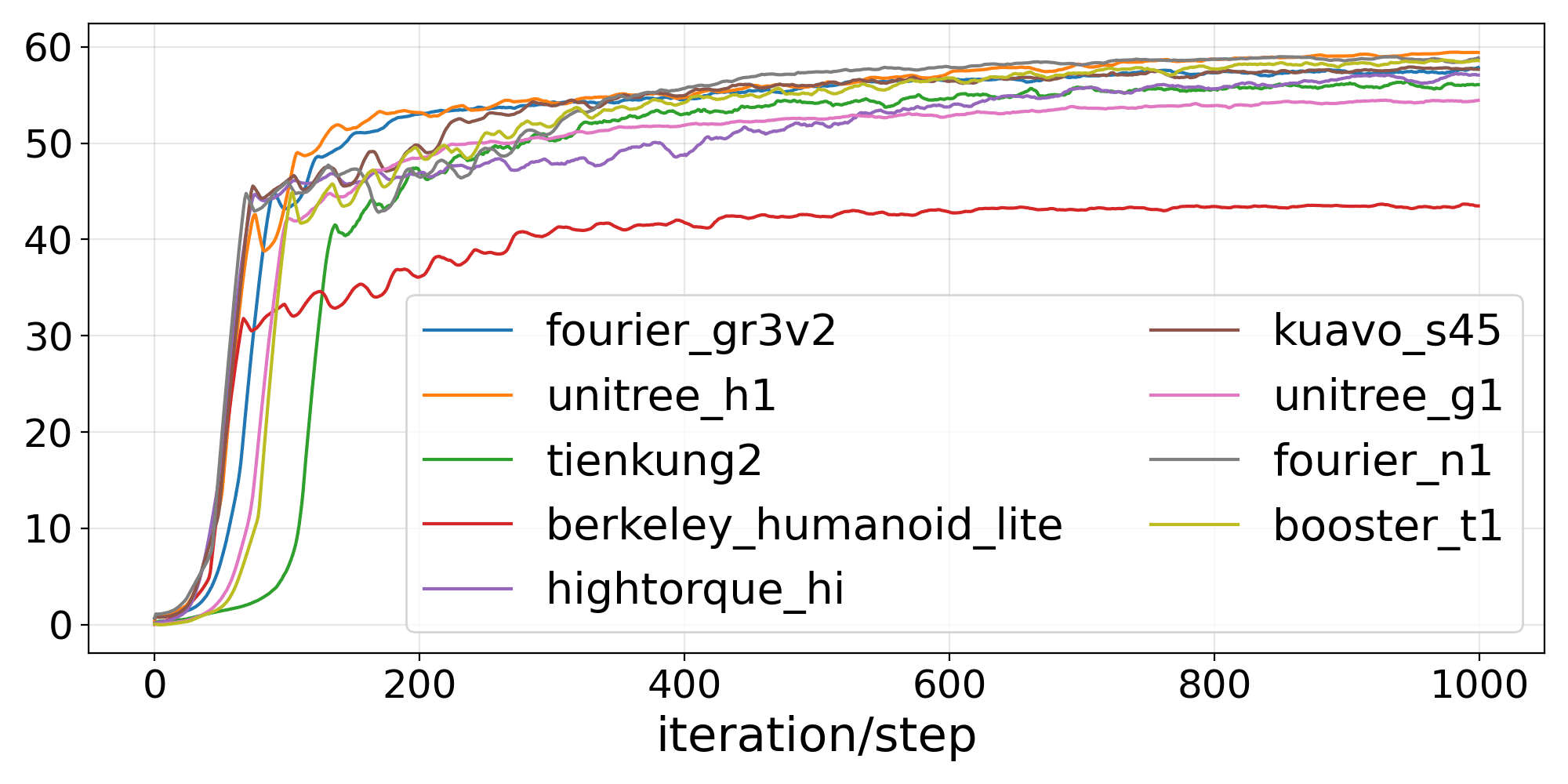}
    \includegraphics[width=0.315\linewidth]{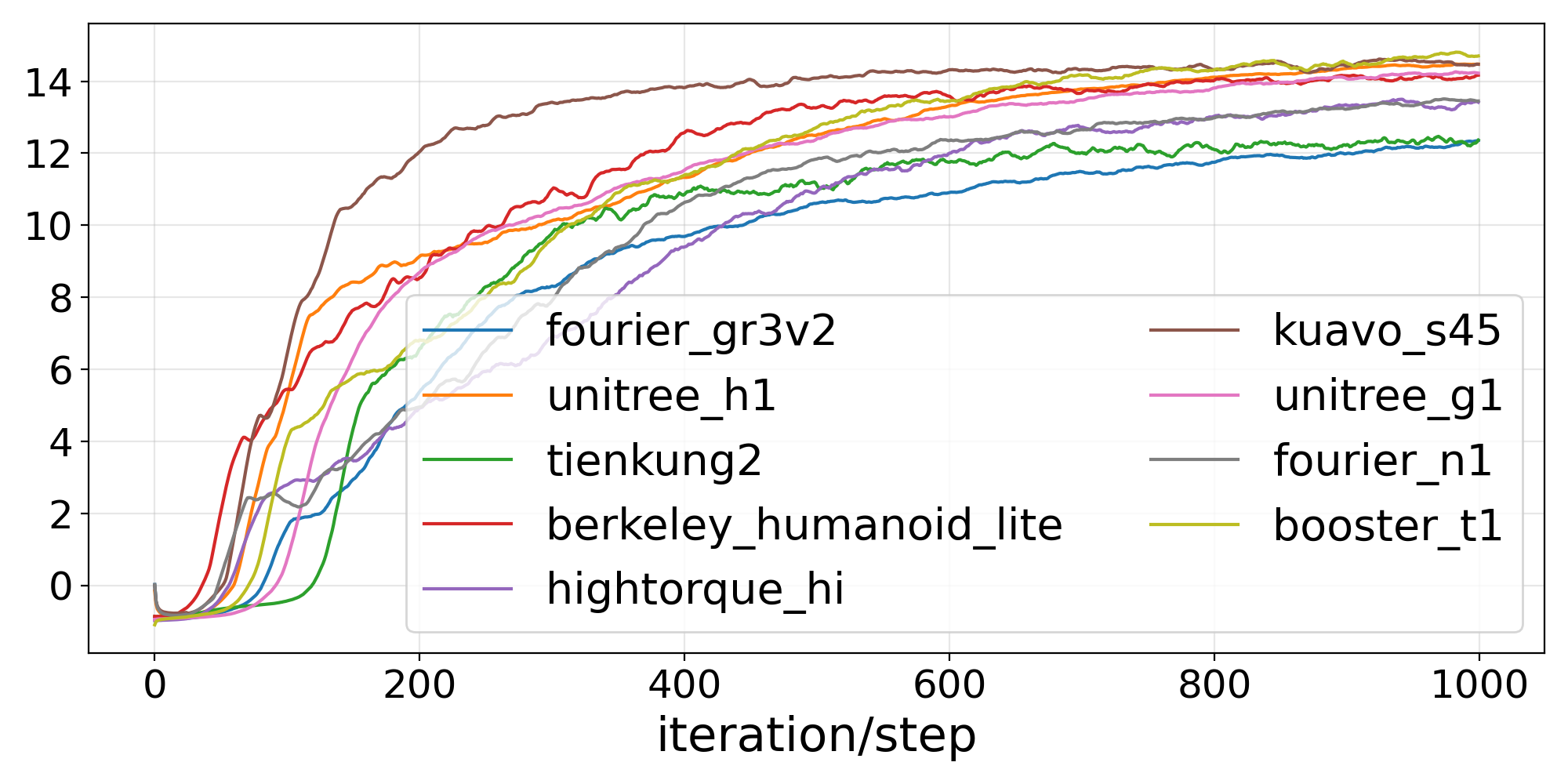}
    \includegraphics[width=0.315\linewidth]{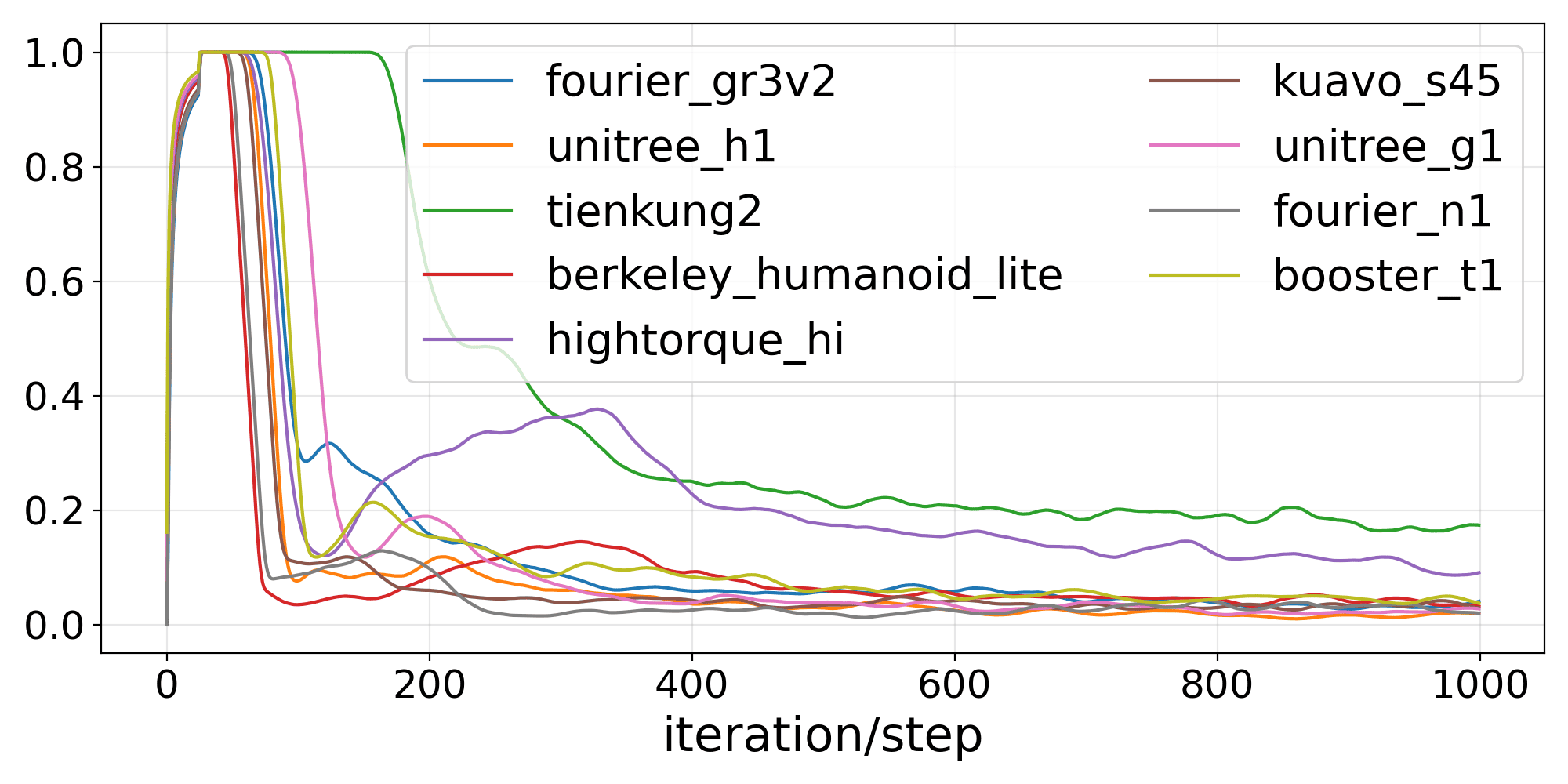}
    \includegraphics[width=0.315\linewidth]{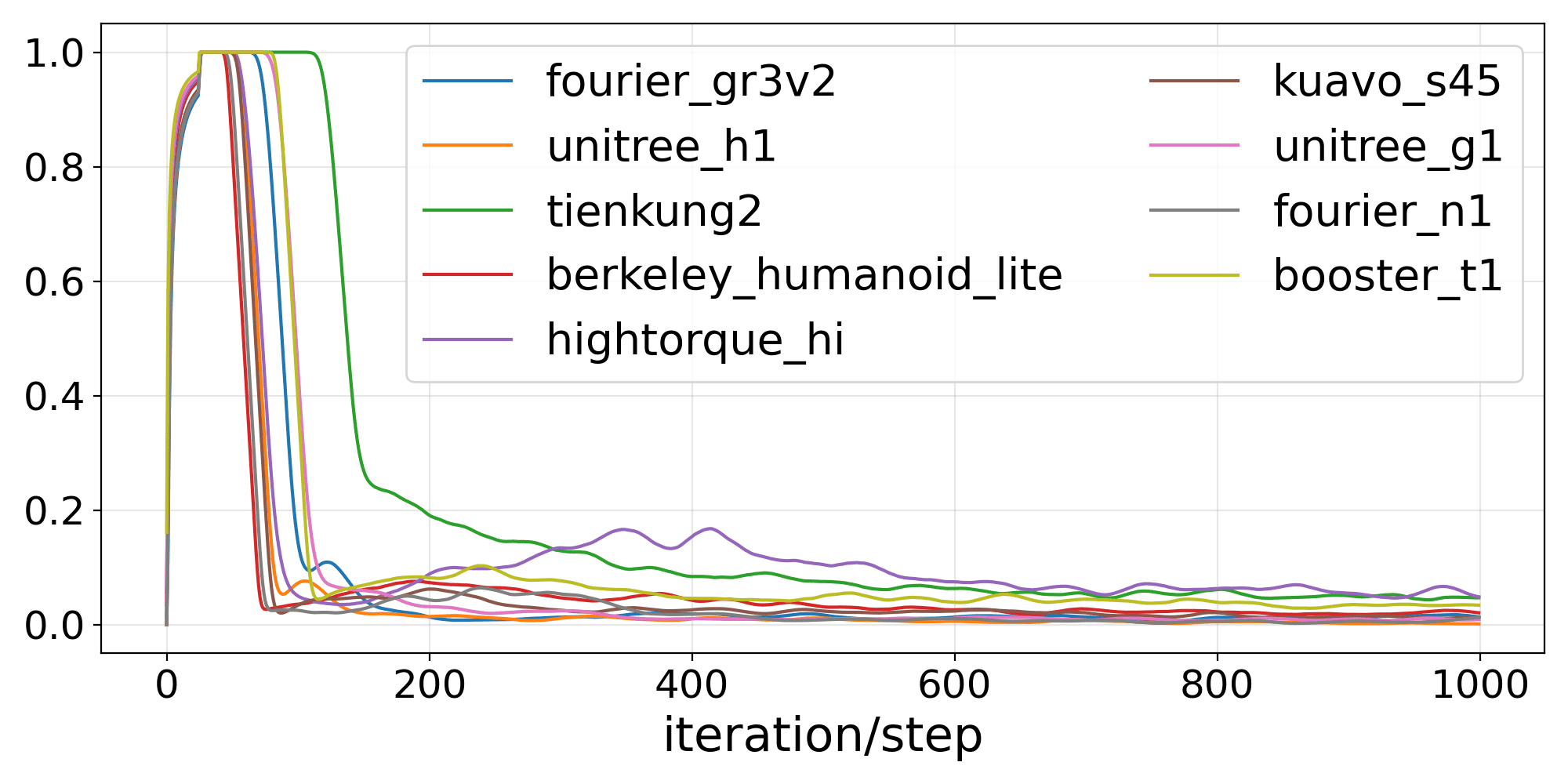}
    \includegraphics[width=0.315\linewidth]{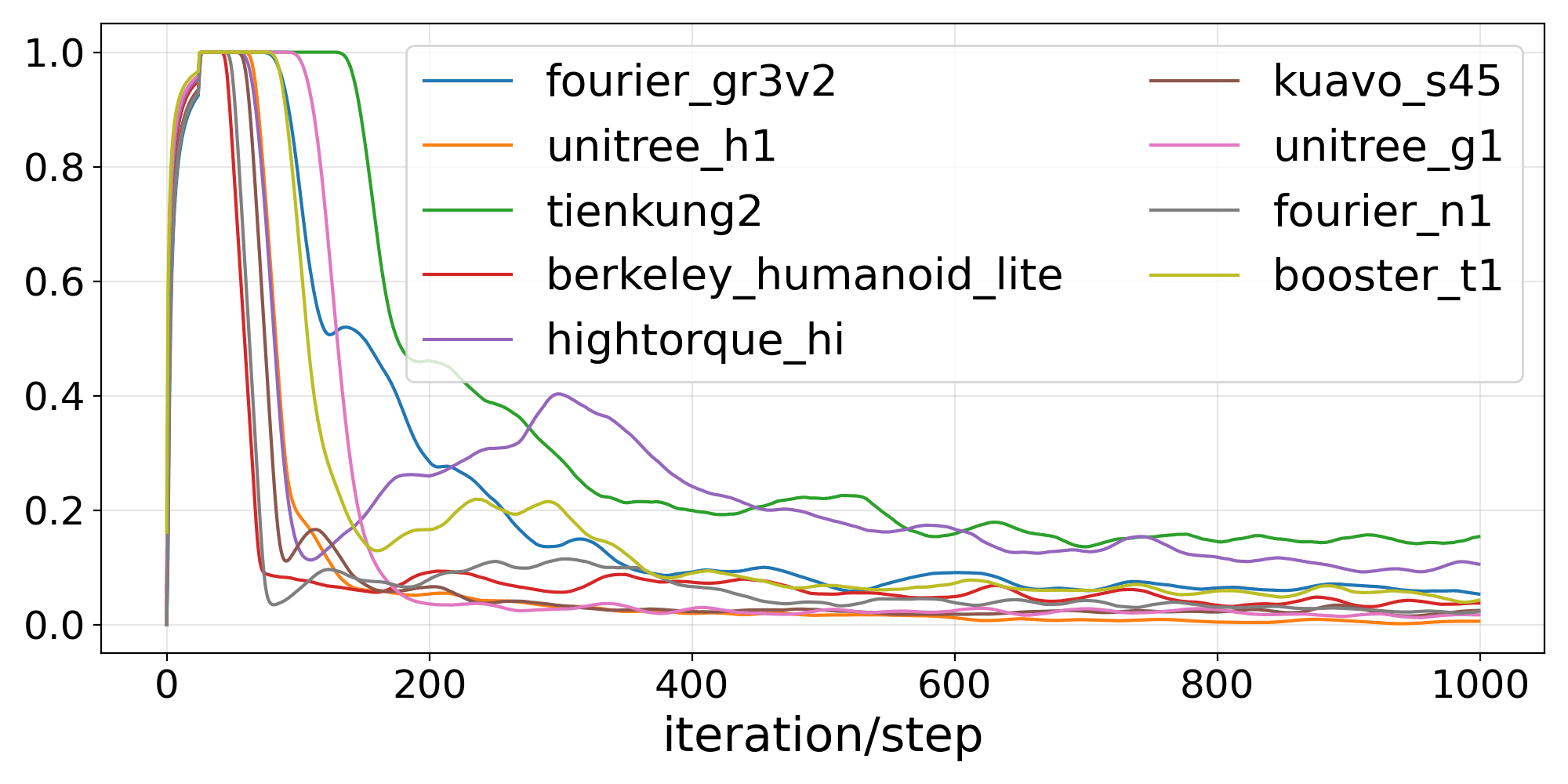}
    \includegraphics[width=0.315\linewidth]{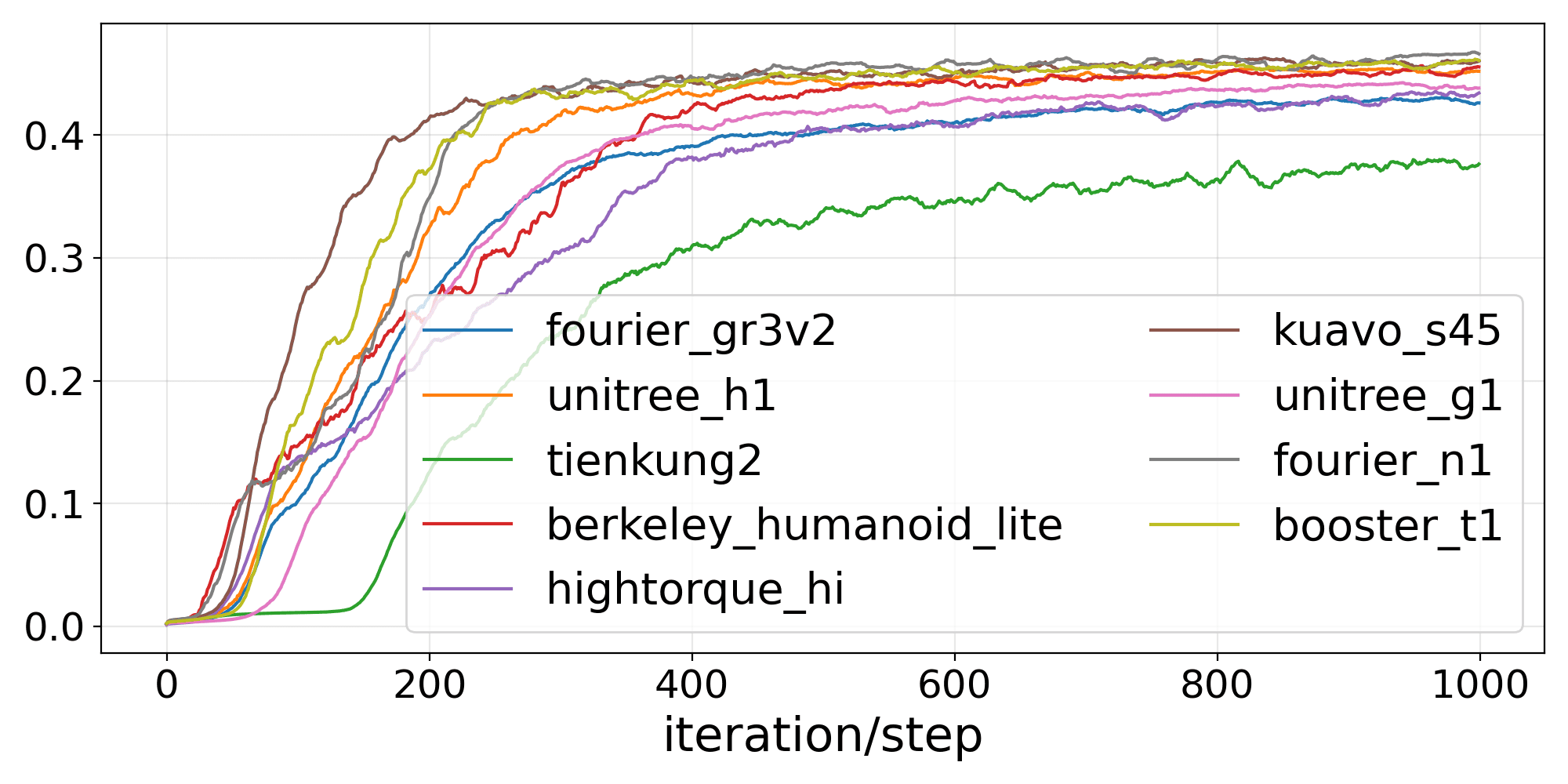}
    \includegraphics[width=0.315\linewidth]{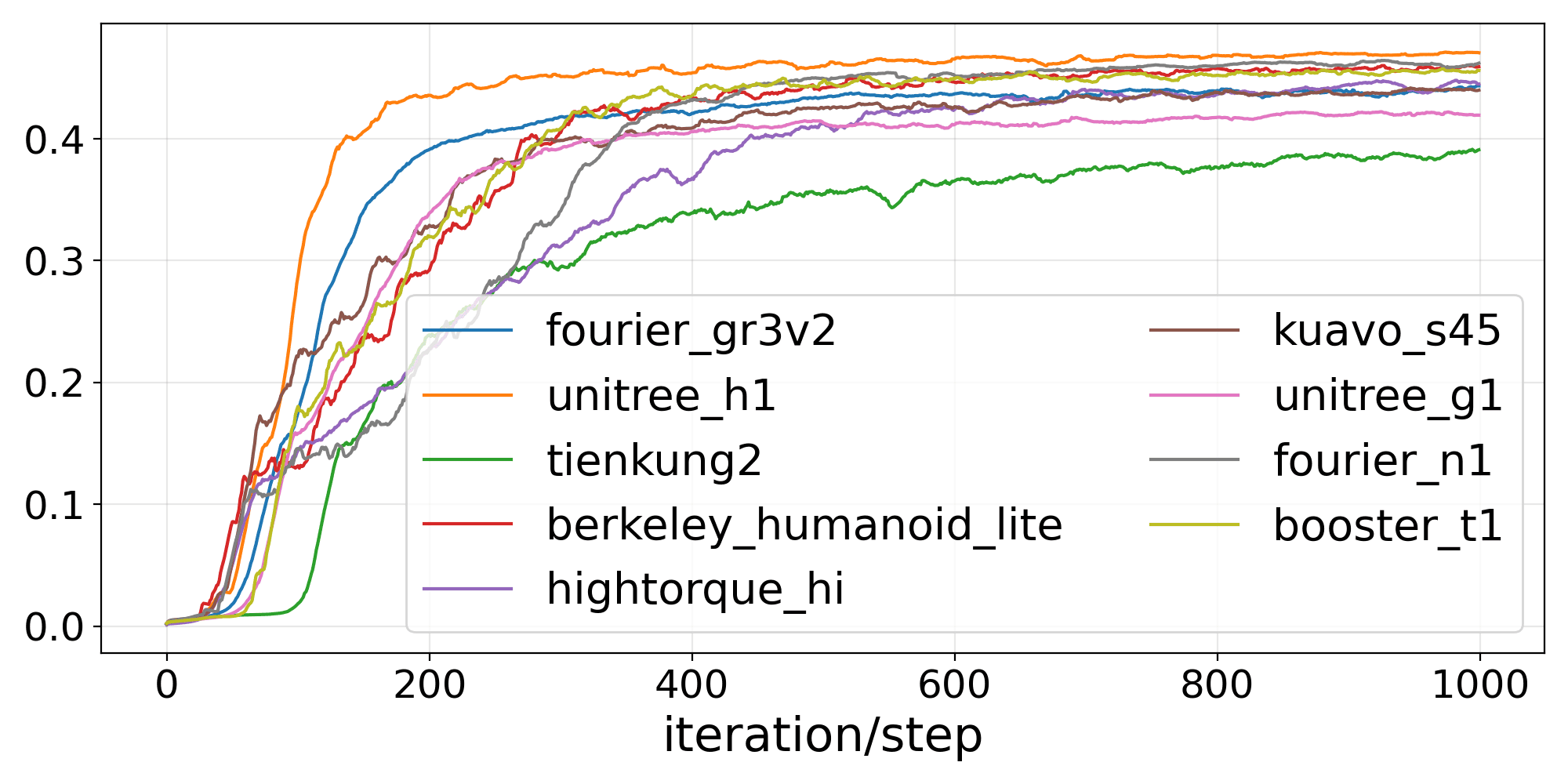}
    \includegraphics[width=0.315\linewidth]{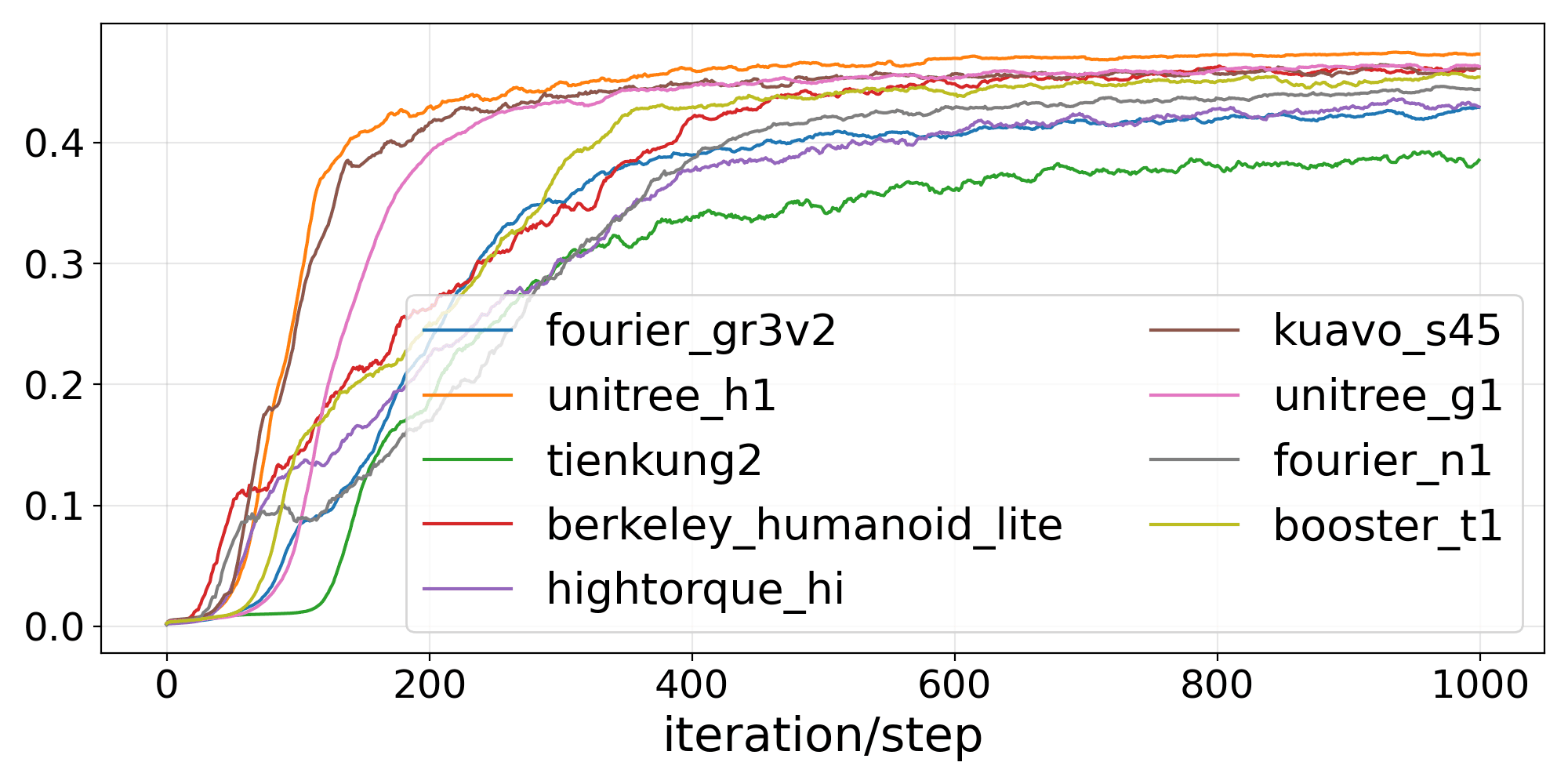}
    \text{a.}\includegraphics[width=0.31\linewidth]{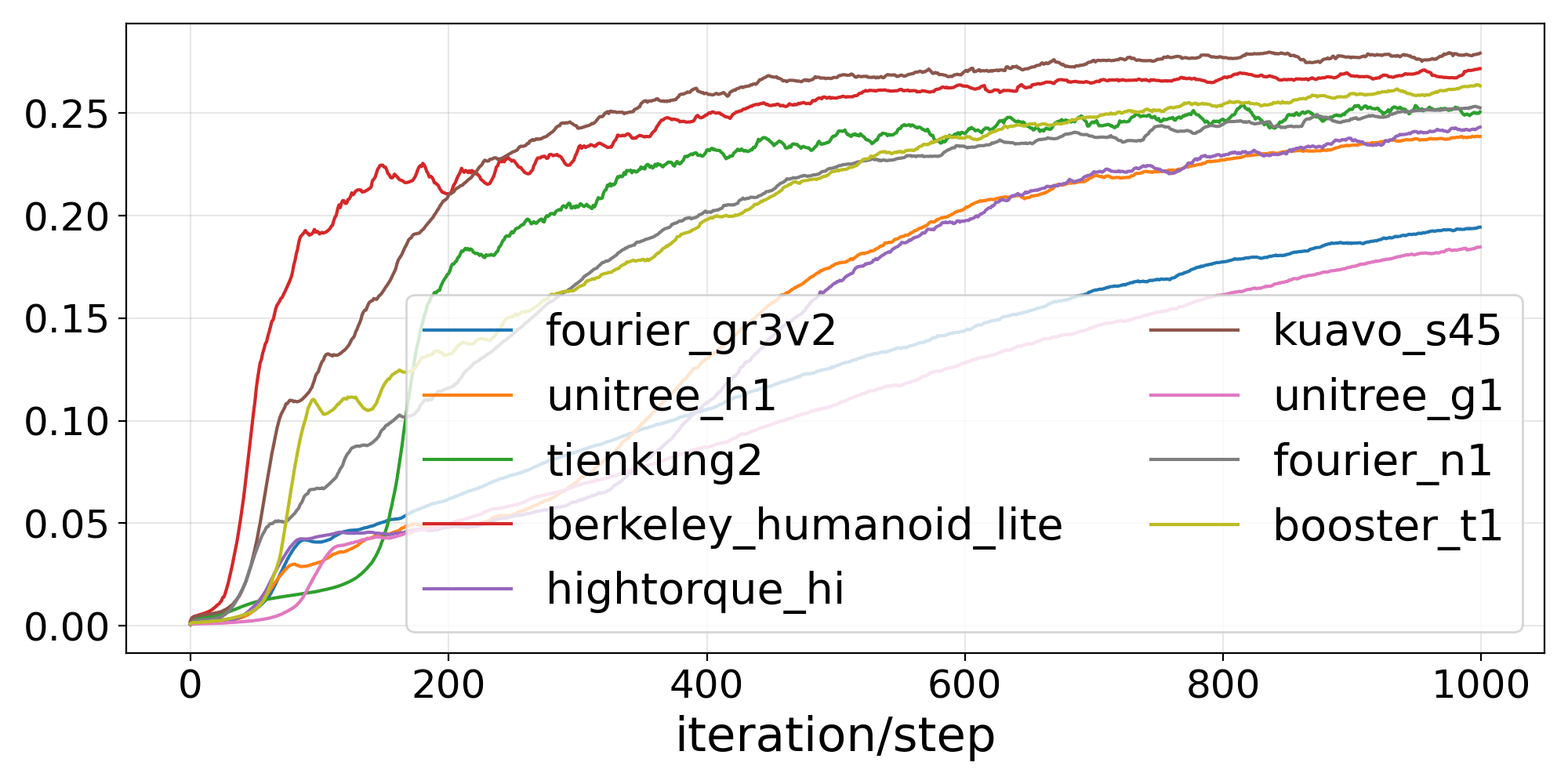}
    \text{b.}\includegraphics[width=0.31\linewidth]{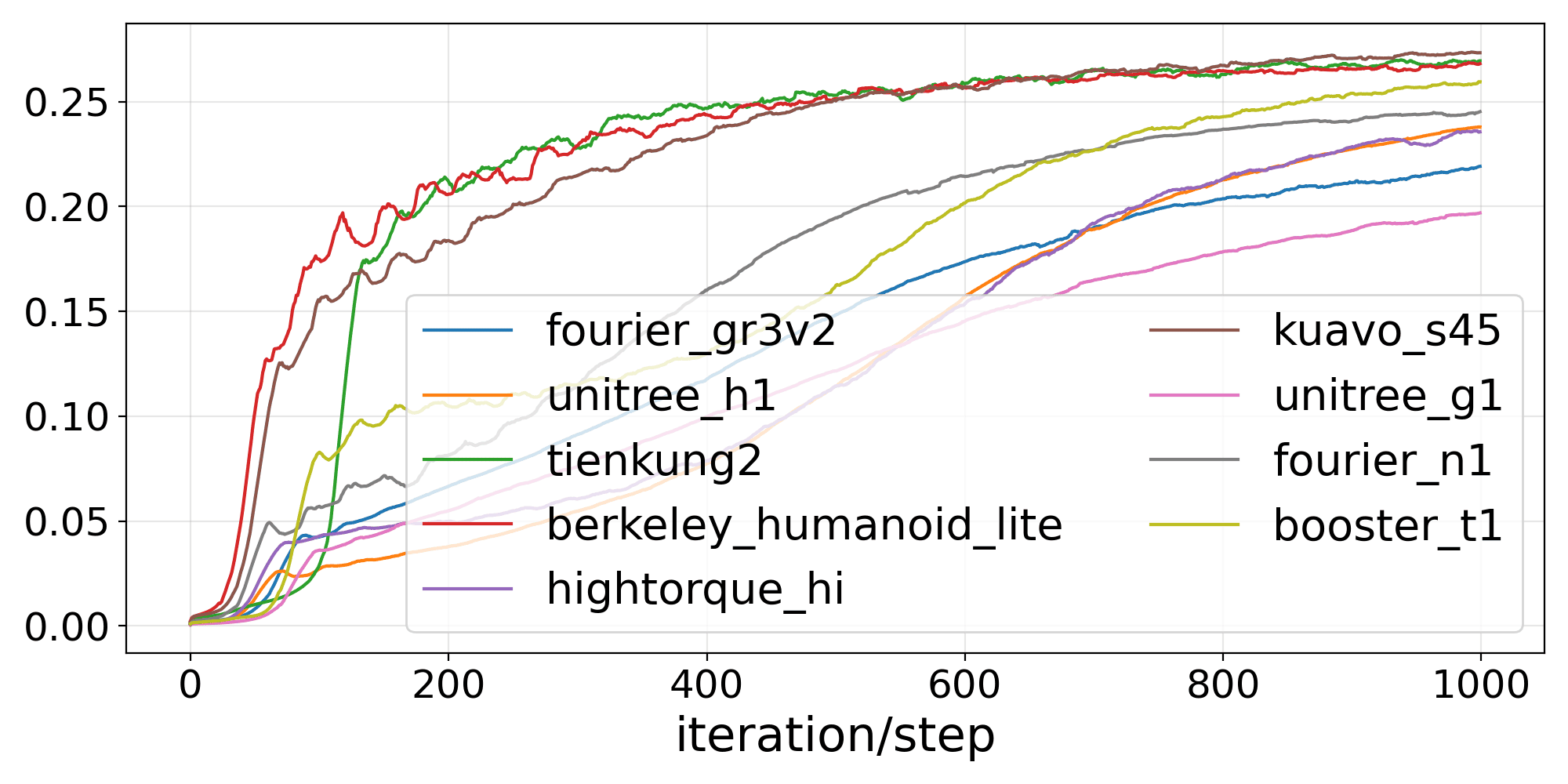}
    \text{c.}\includegraphics[width=0.31\linewidth]{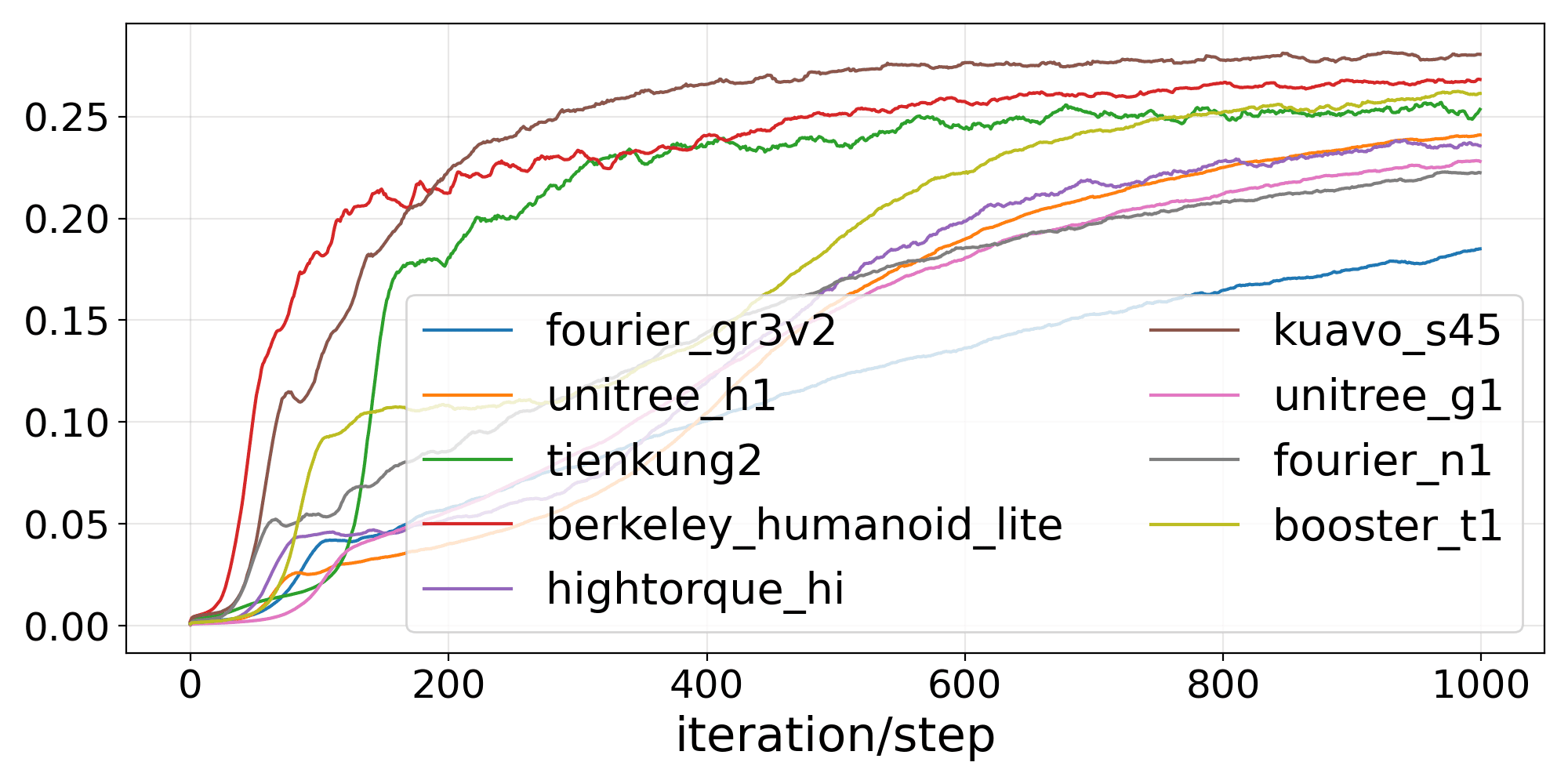}
    \caption{Benchmark comparison of the Baseline, DRS, and Multi-Potential PBRS in terms of total reward, fall rate, tracking reward in the $x$--$y$ velocity, and tracking reward in angular velocity. The first column shows (a) Baseline, (b) Direct Reward Shaping (DRS), and (c) Multi-Potential PBRS. From top to bottom, the rows show total reward, fall rate, tracking reward for $x$--$y$ velocity, and tracking reward for angular velocity.}
    \label{benchmarkcompare}
\end{figure*}
\begin{figure} 
    \centering
    \includegraphics[width=0.49\linewidth]{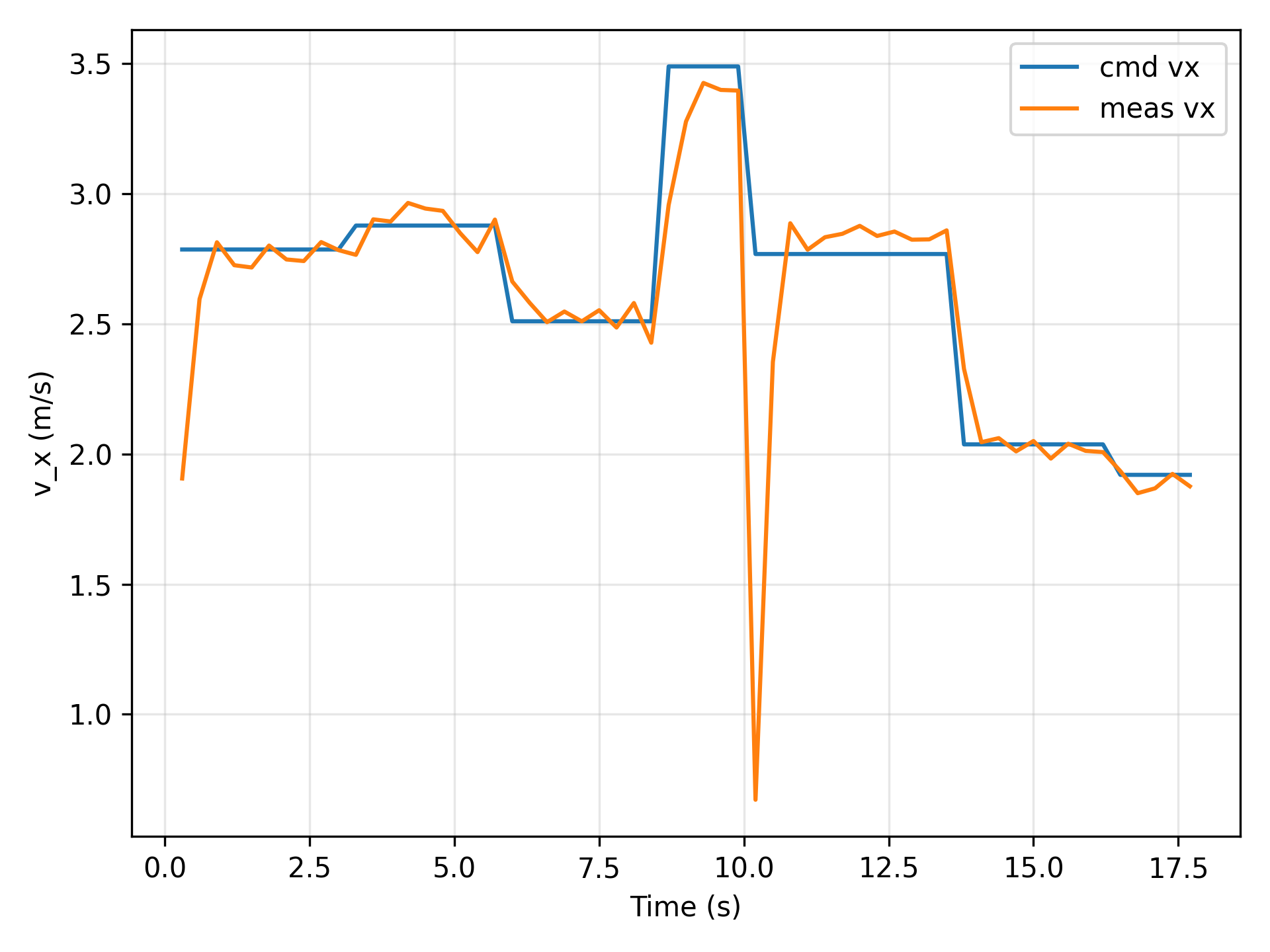}
    \includegraphics[width=0.49\linewidth]{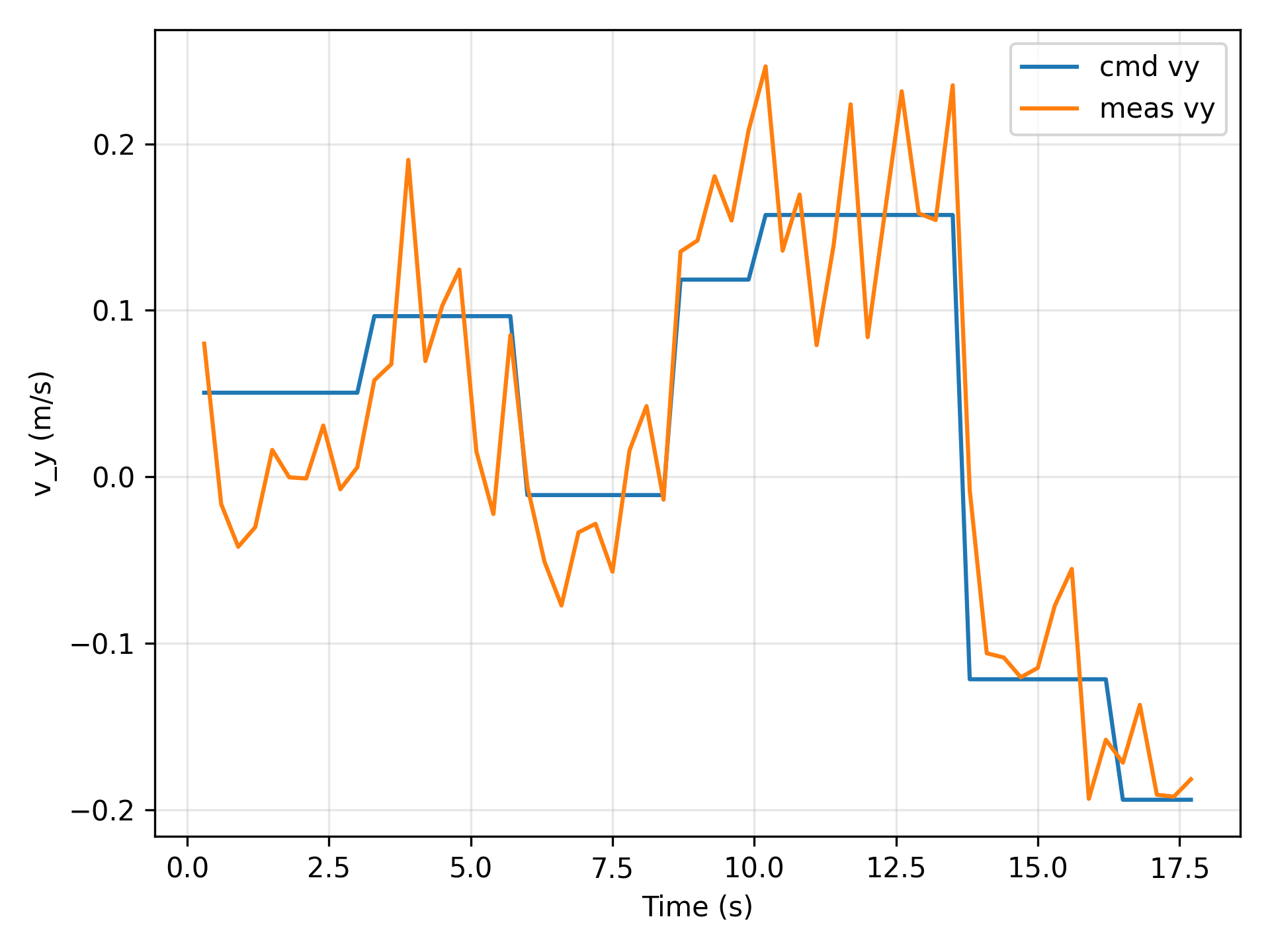}
    \includegraphics[width=0.9\linewidth]{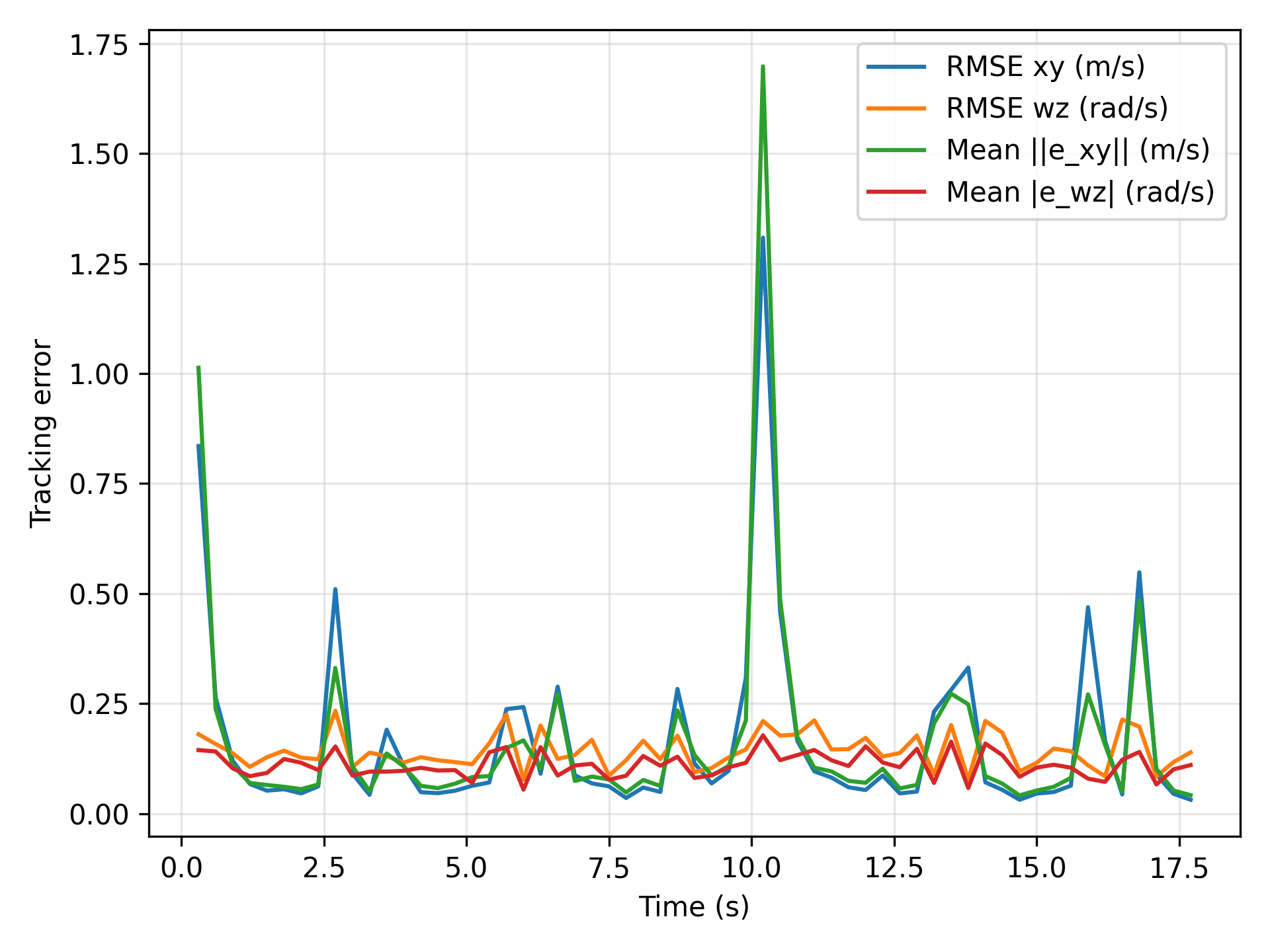}
    \caption{Velocity tracking plots for the policy running on the Fourier N1 robot over a 20-second interval. The robot was respawned every 10 seconds, as reflected by the jumps in the tracking error.}
    \label{trackingabilityn1}
\end{figure}
\subsection{Implementation Details}
The locomotion task is implemented in NVIDIA IsaacLab using the PPO-Clip algorithm with RSL-RL \cite{schwarke2025rsl}. All hyperparameters are listed in Table~\ref{tab:train_hyperparams}. The task is trained on a computer equipped with an Intel i7-9700F CPU and an NVIDIA GeForce RTX 2060 GPU (6 GB). The simulation runs at 200 Hz, and the control frequency is 100 Hz. Each training run uses 2048 agents and lasts for 1000 interaction steps. The training takes approximately 50 minutes to complete.

\section{RESULTS}
\begin{figure*} [!ht]
    \centering
    \includegraphics[width=0.315\linewidth]{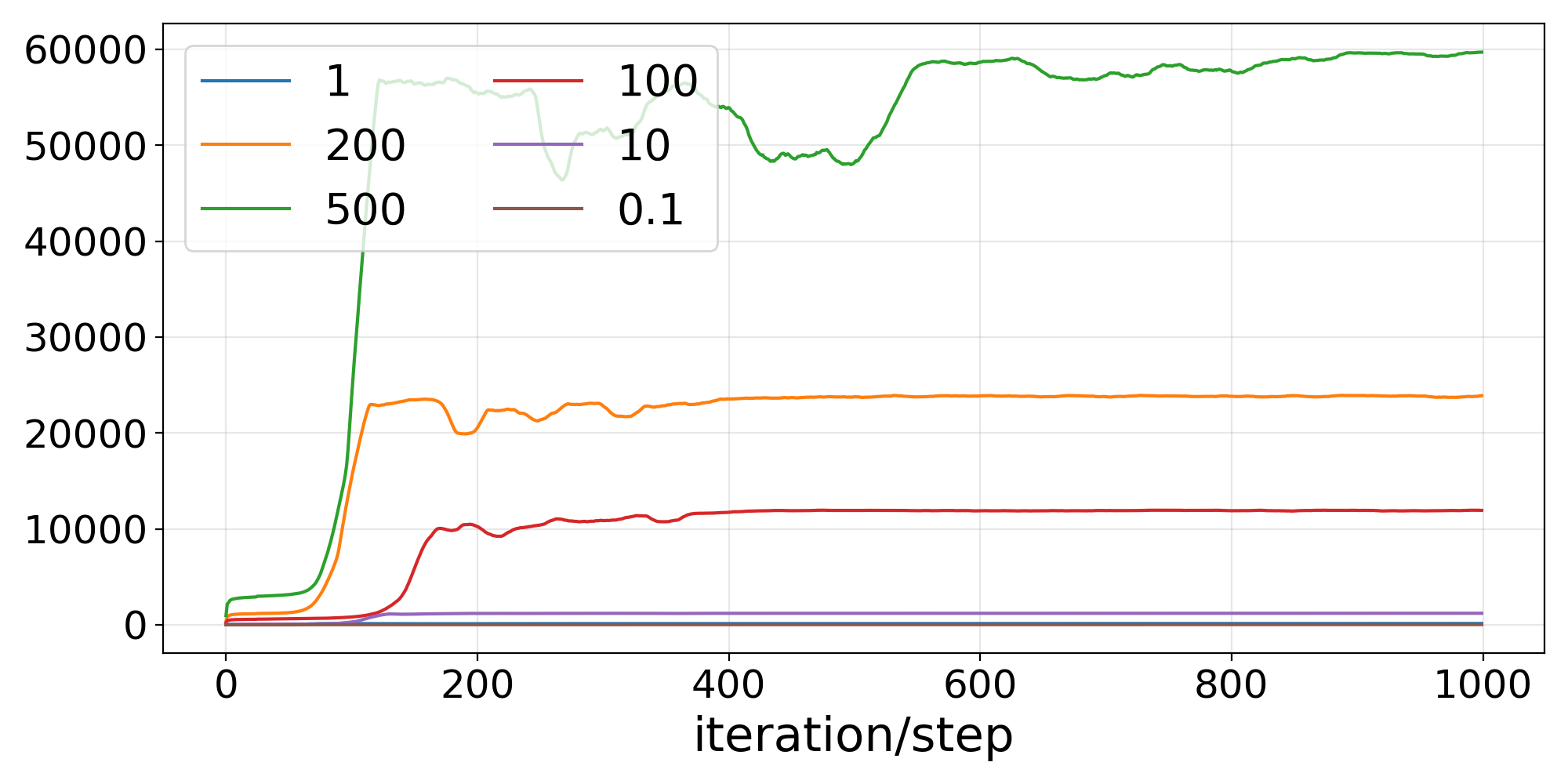}
    \includegraphics[width=0.315\linewidth]{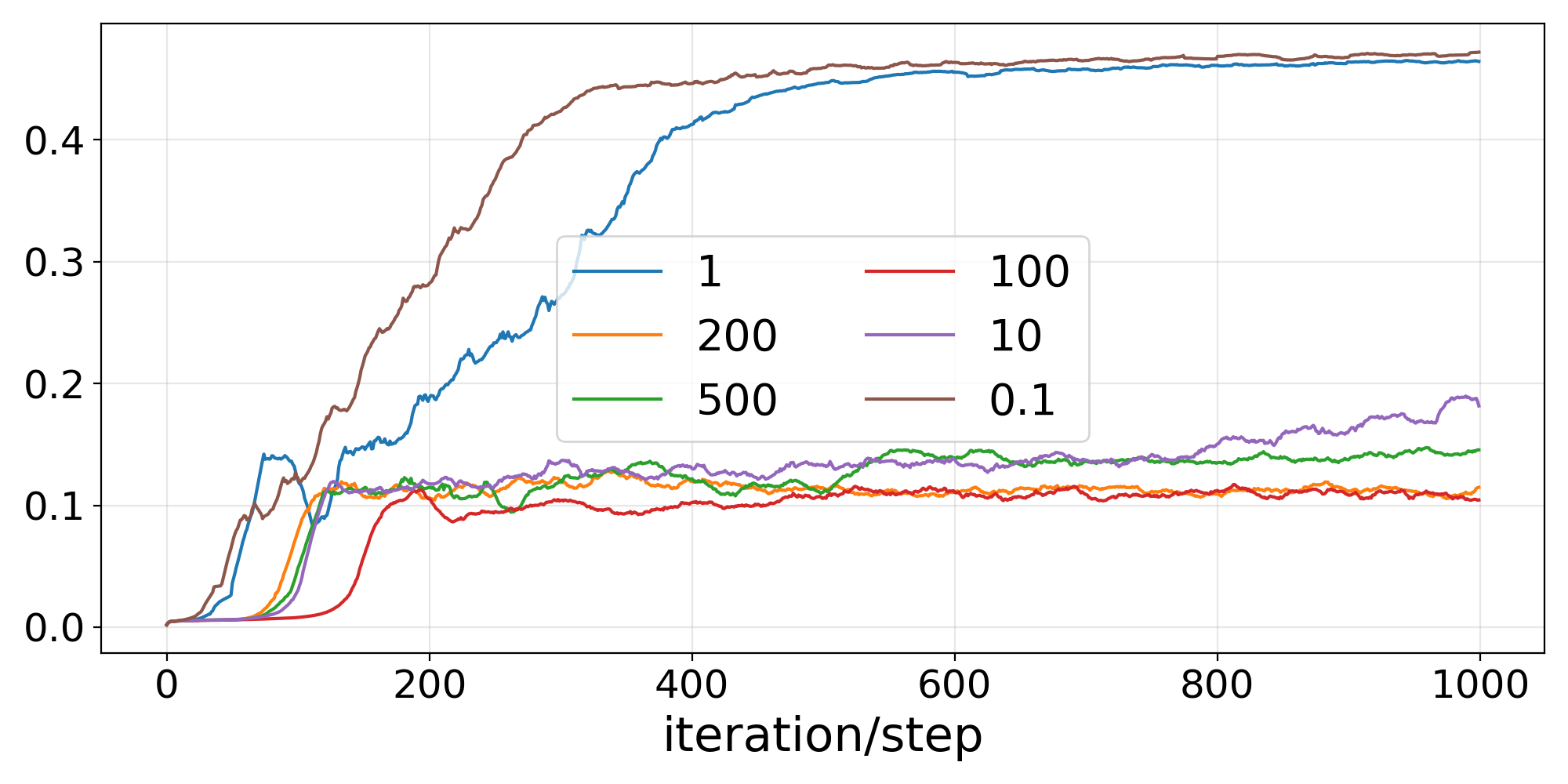}
    \includegraphics[width=0.315\linewidth]{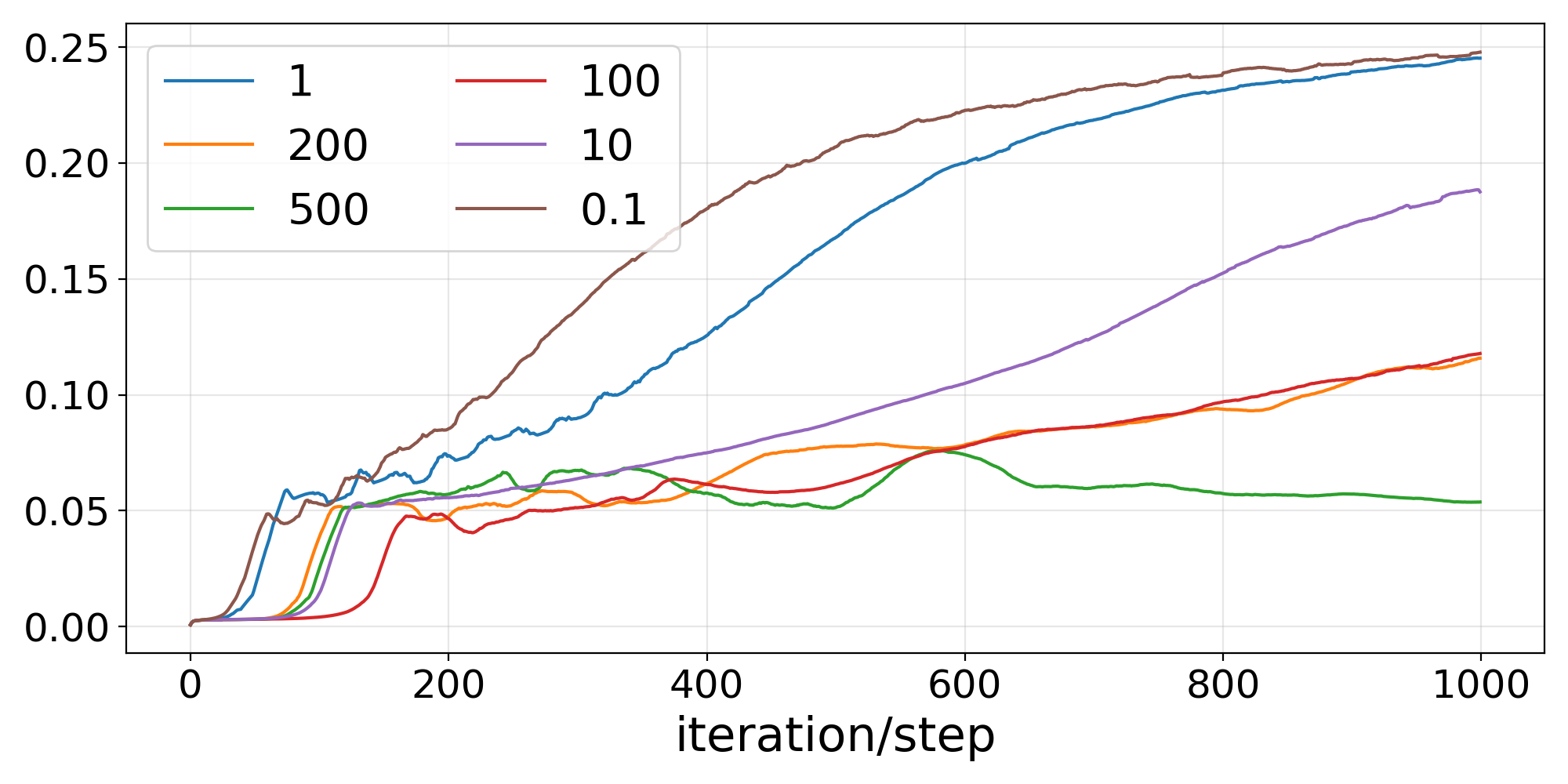}
    \text{a.}\includegraphics[width=0.31\linewidth]{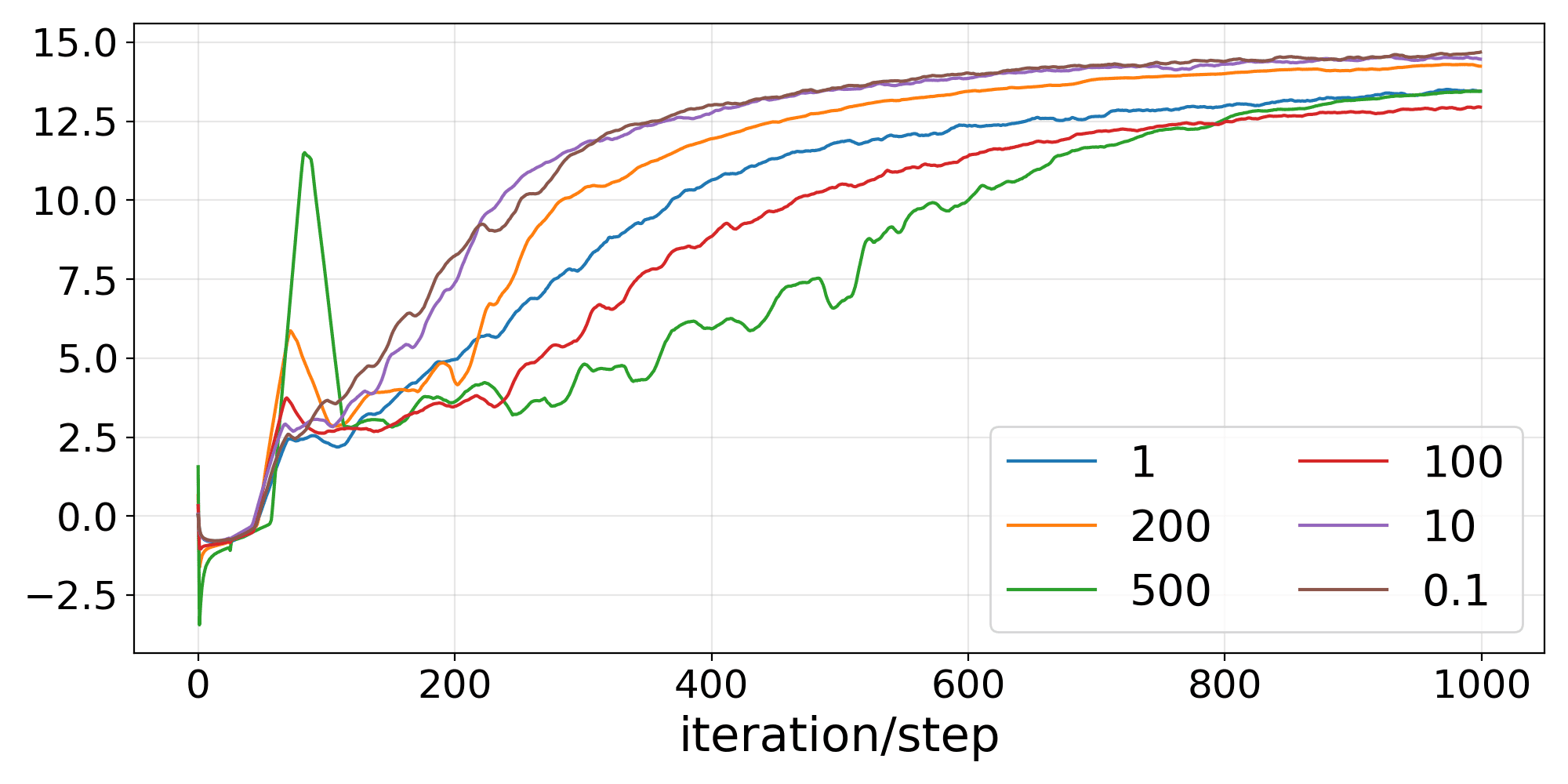}
    \text{b.}\includegraphics[width=0.31\linewidth]{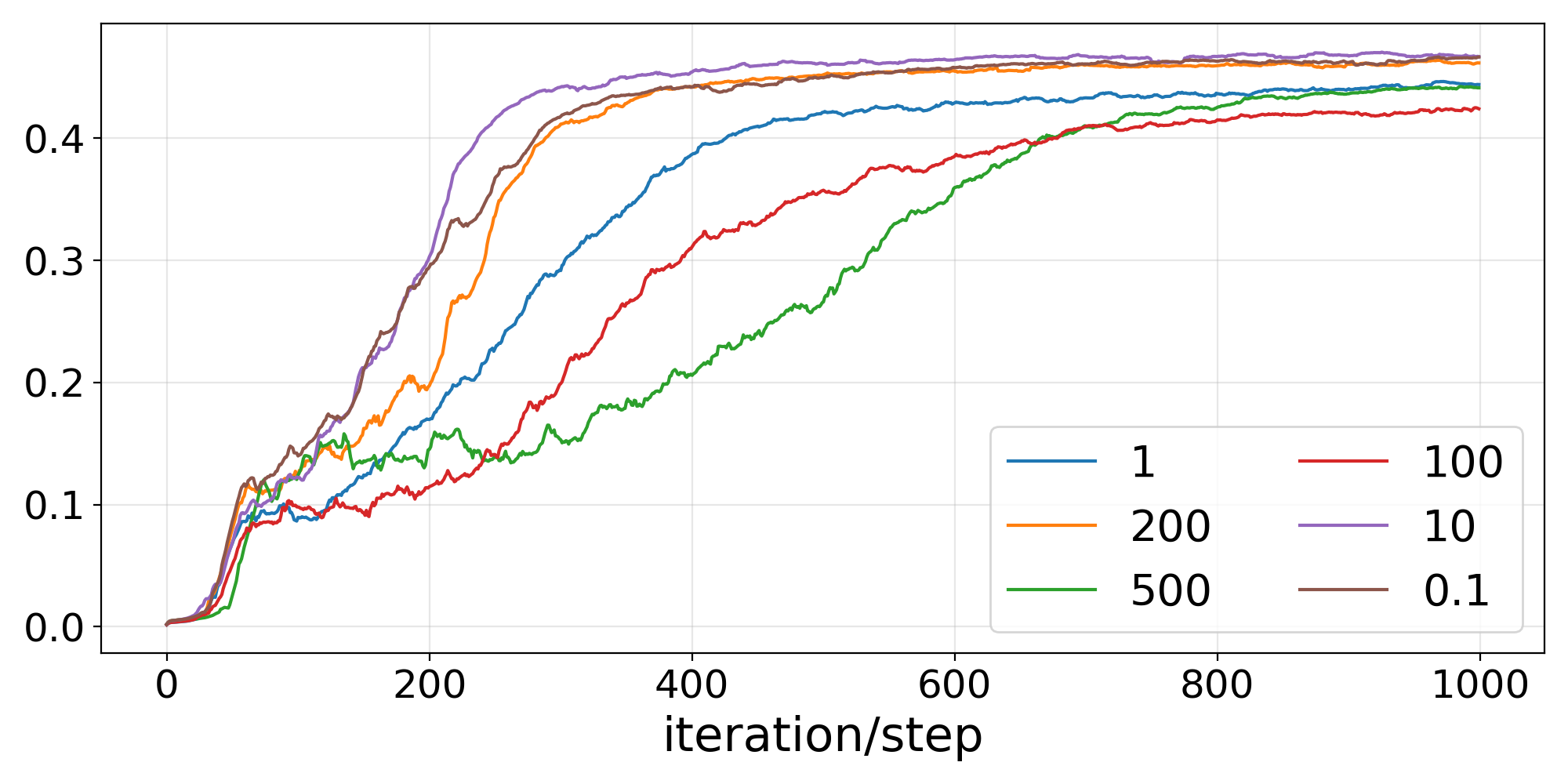}
    \text{c.}\includegraphics[width=0.31\linewidth]{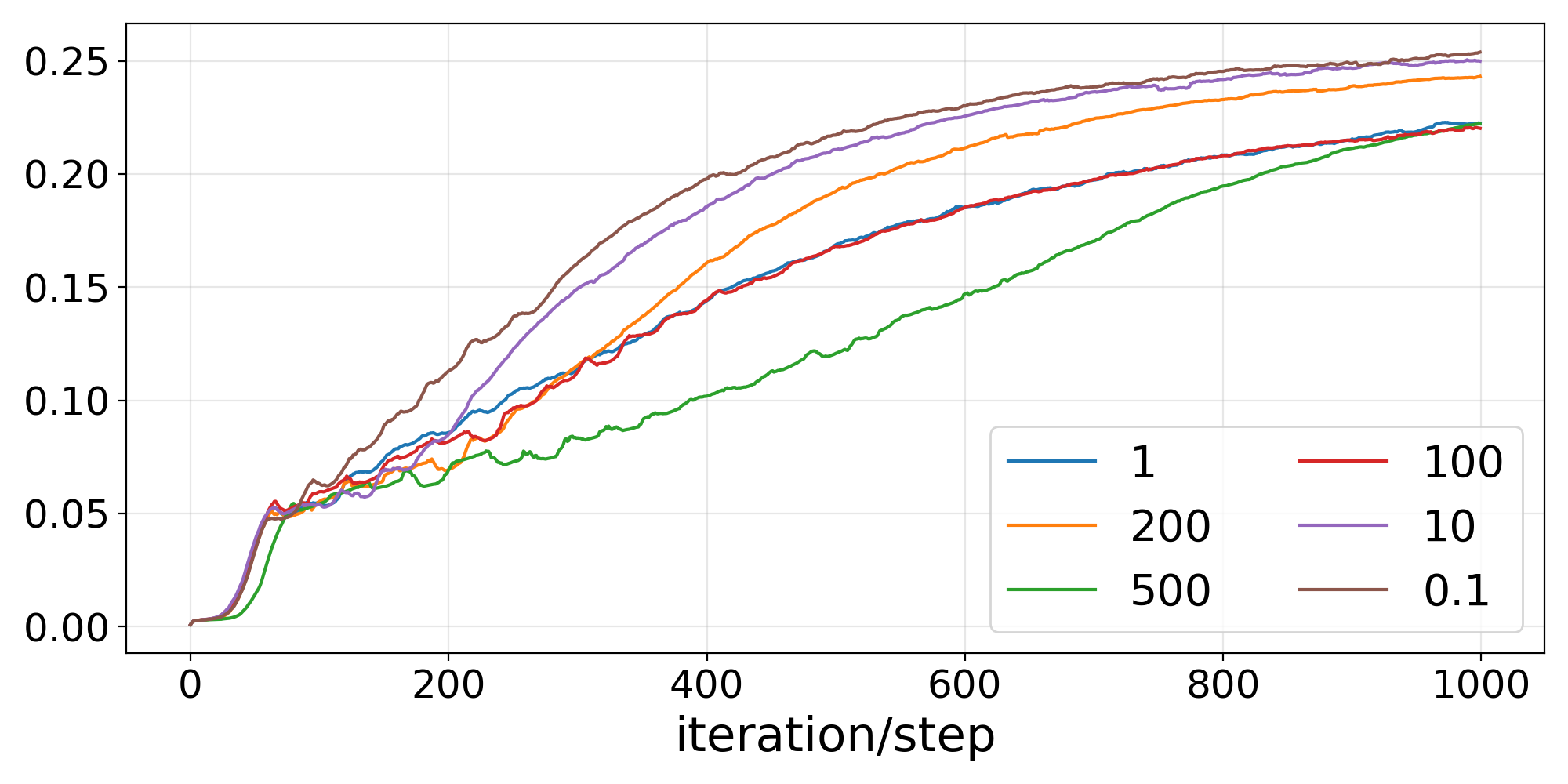}
    \vspace{-10pt}
    \caption{To compare the robustness of potential-based shaping against DRS, we train policies using DRS and MPBRS while varying the shaping weight from 0.1 to 500. The first row shows DRS, and the second row shows MPBRS. The comparison is based on (a) total reward, (b) tracking reward in the $x$--$y$ direction, and (c) tracking reward for yaw rate.}
    \vspace{-10pt}
    \label{benchmarkweightsensitive}
\end{figure*}
\begin{figure*} [!ht]
    \centering
    \text{a.}\includegraphics[width=0.31\linewidth]{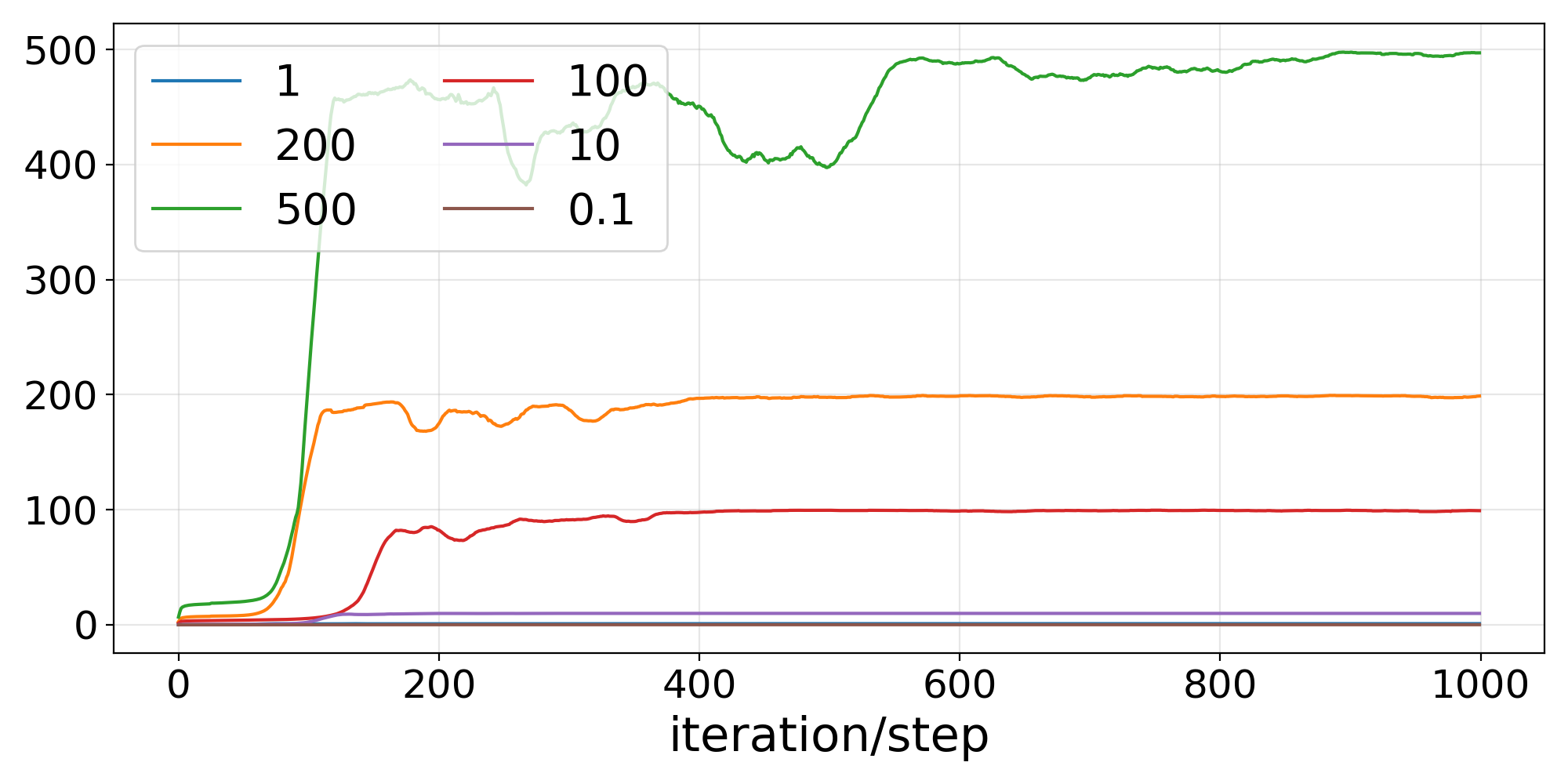}
    \includegraphics[width=0.31\linewidth]{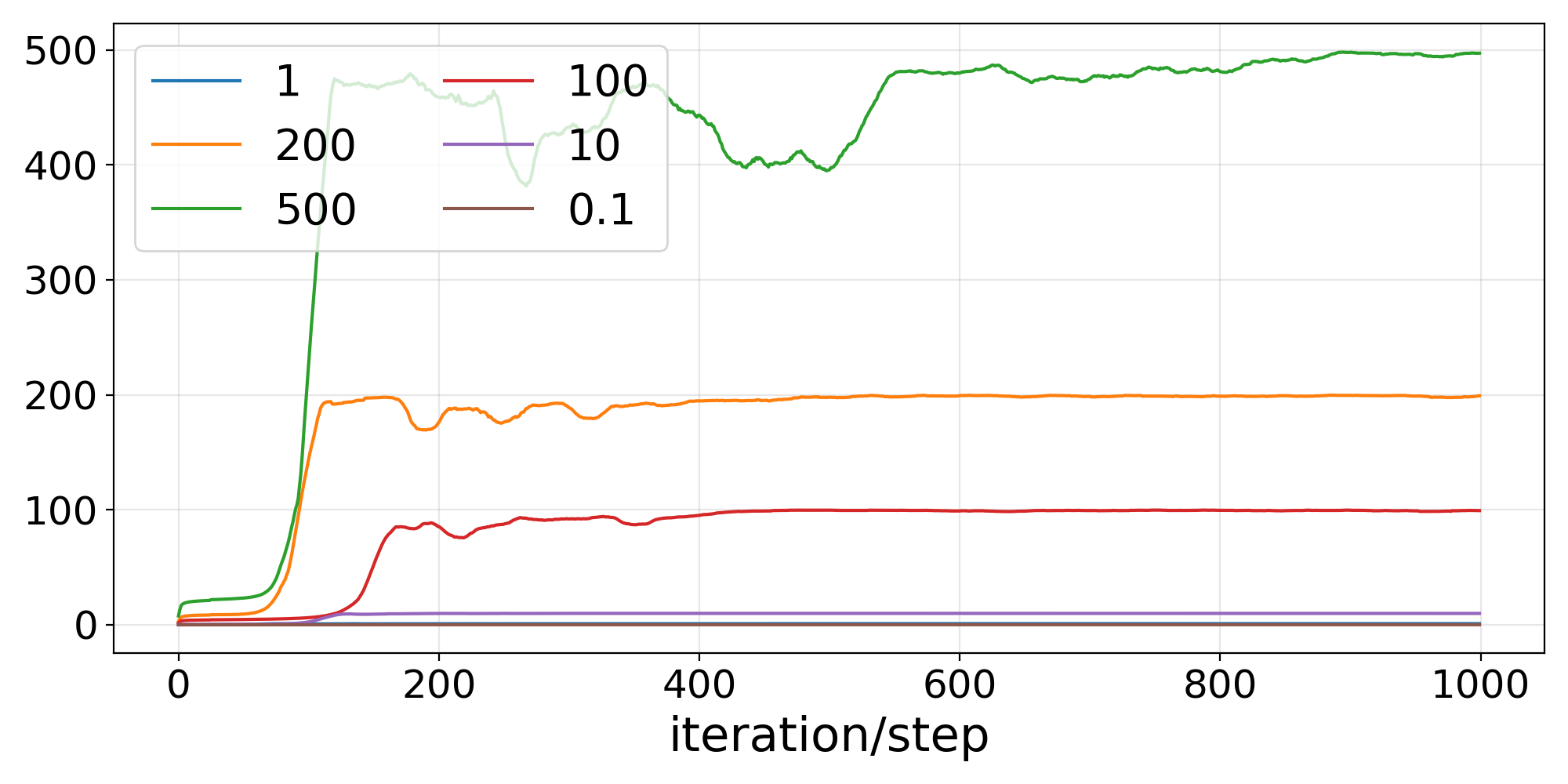}
    \includegraphics[width=0.31\linewidth]{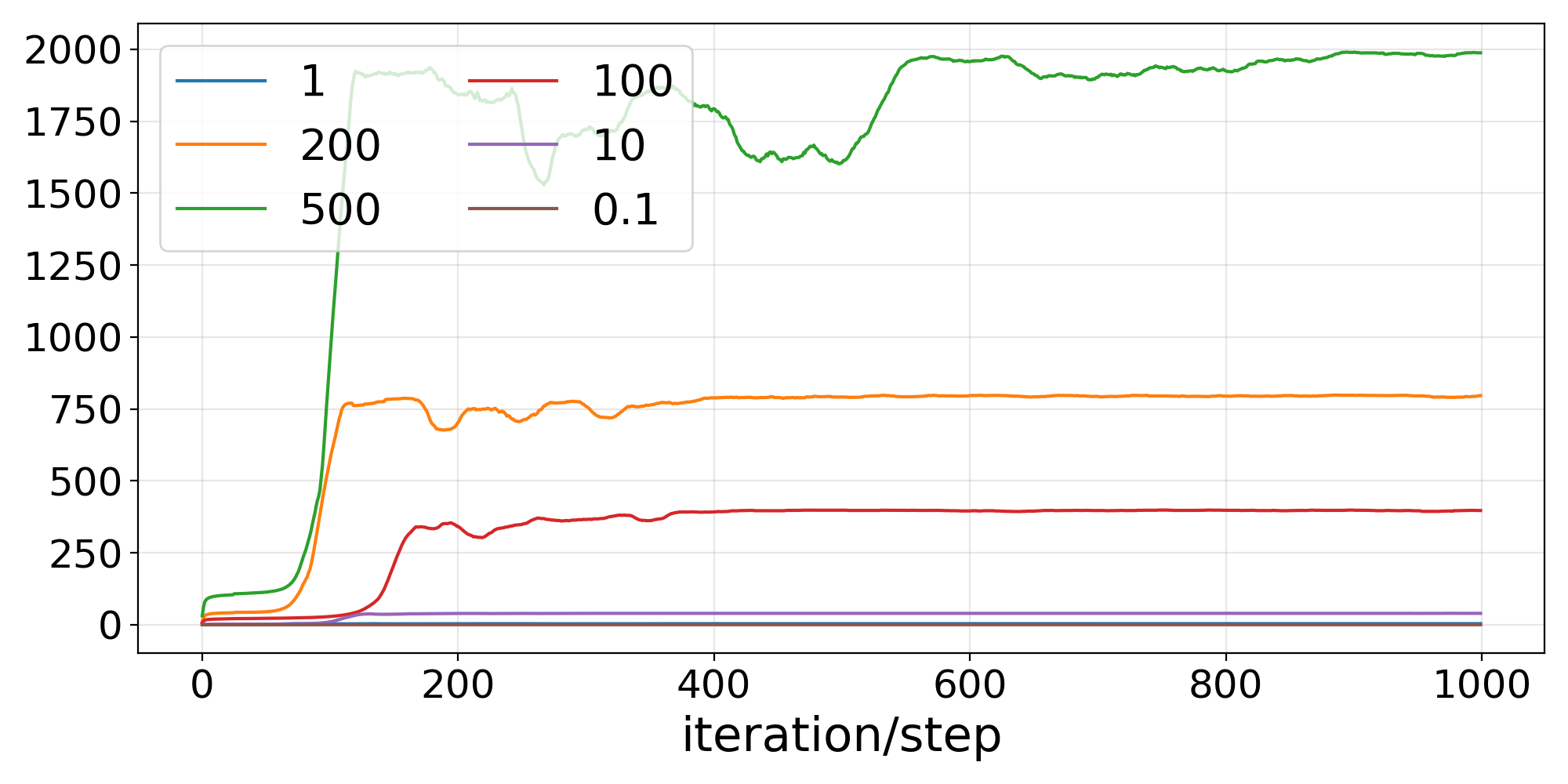}
    \text{b.}\includegraphics[width=0.31\linewidth]{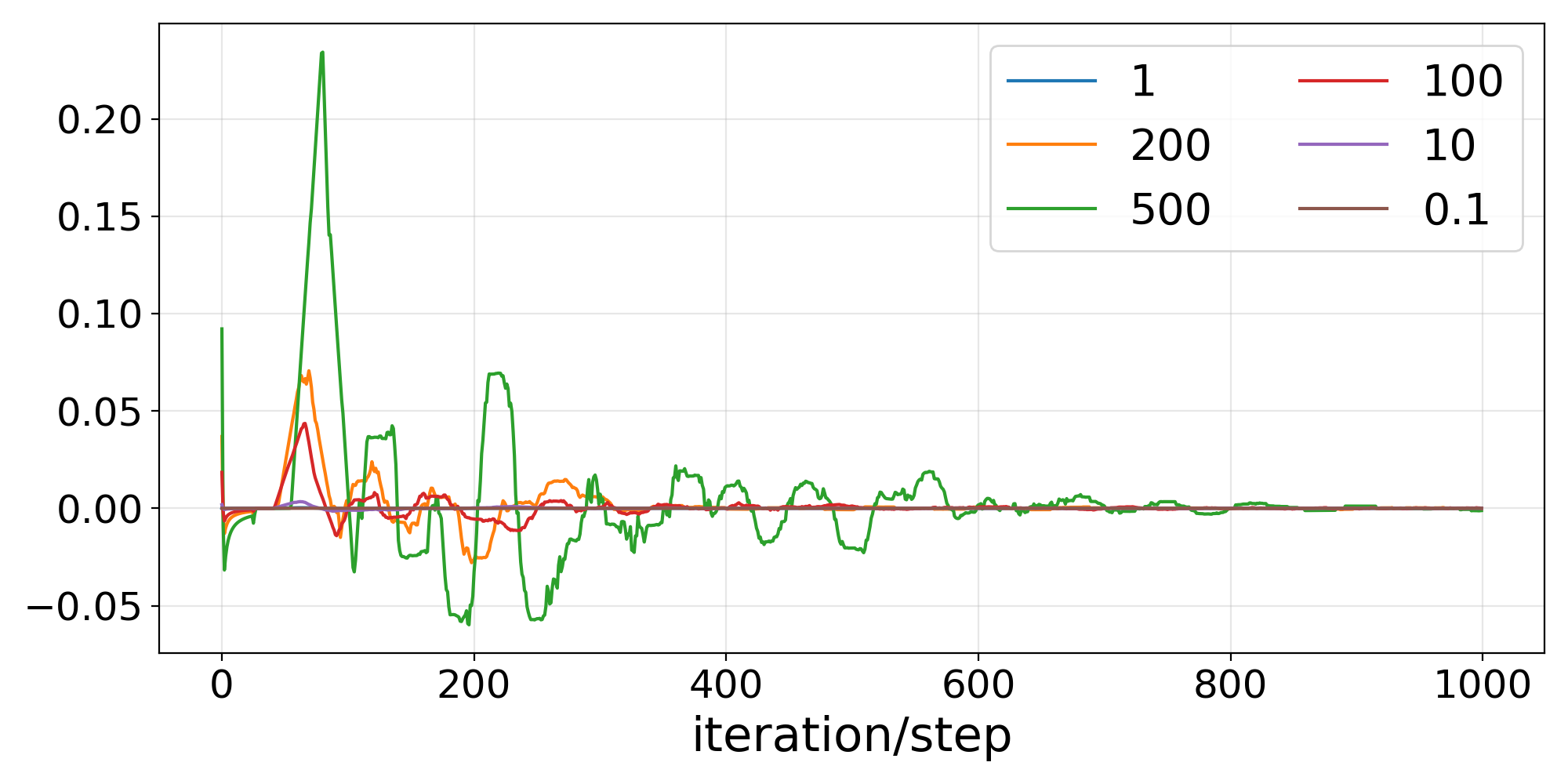}
    \includegraphics[width=0.31\linewidth]{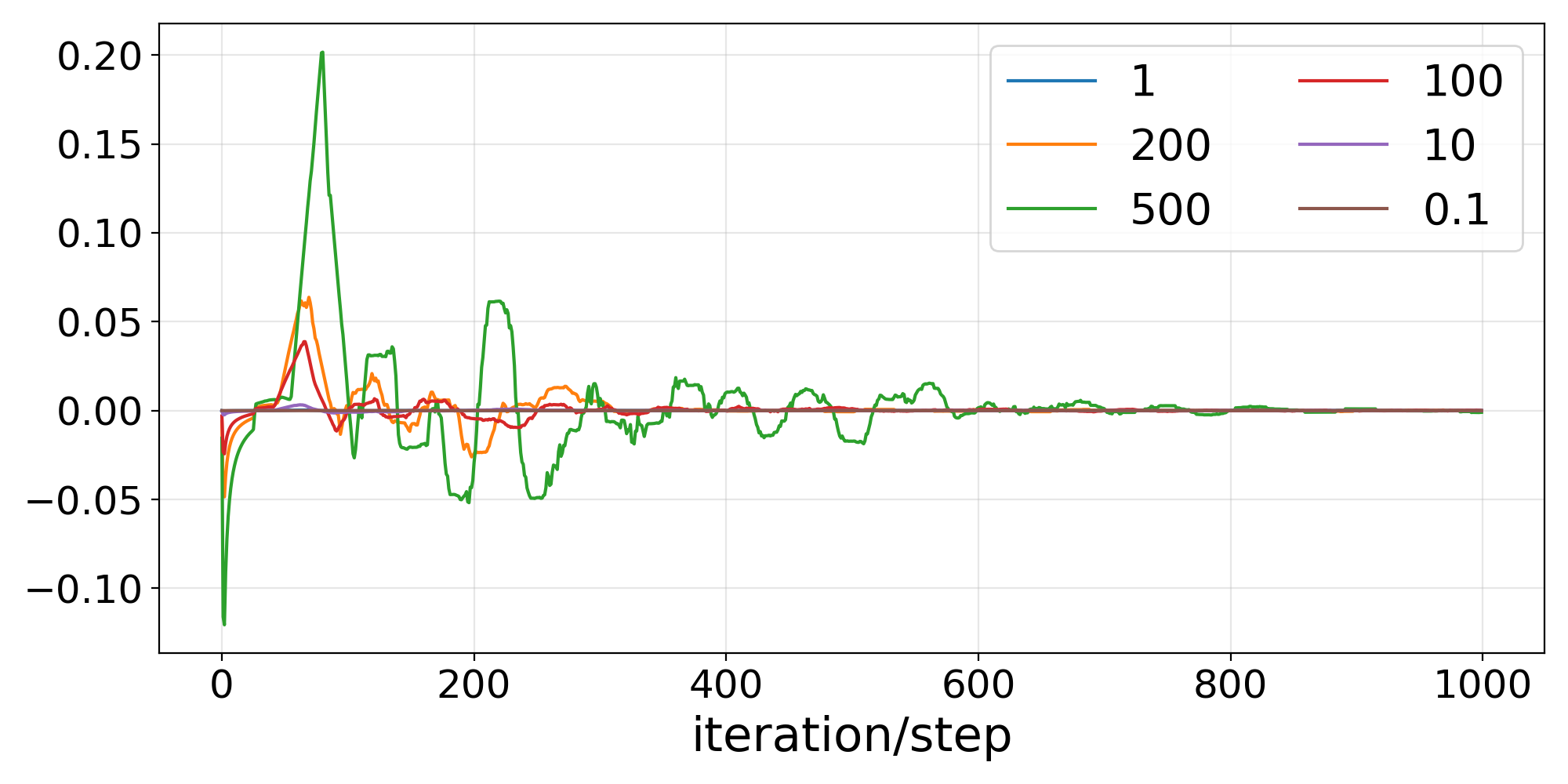}
    \includegraphics[width=0.31\linewidth]{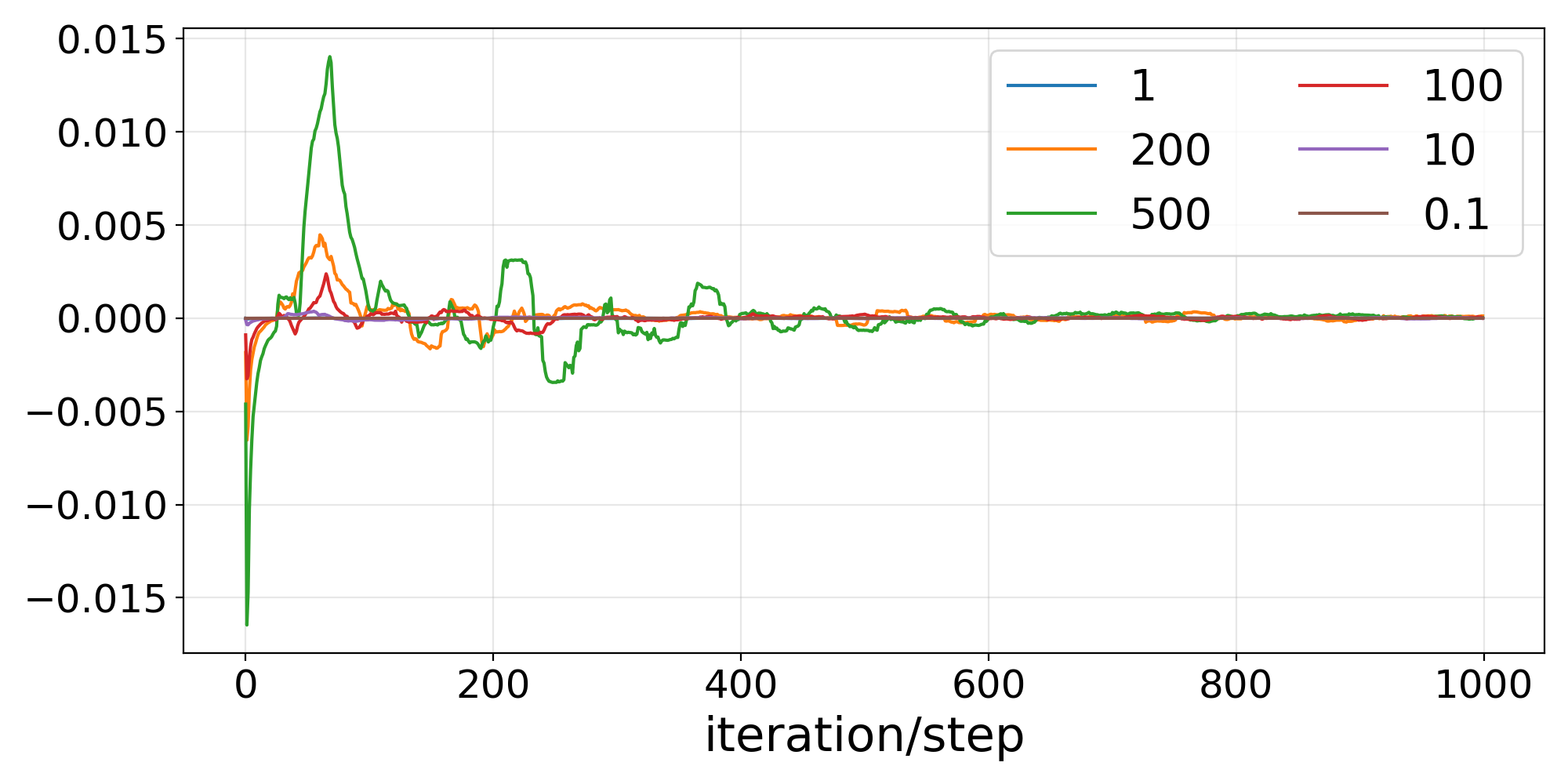}
    \vspace{-10pt}
    \caption{Reward distributions for DRS and the proposed method in the sensitivity-analysis scenario. The first row (a) corresponds to DRS, and the second row (b) corresponds to MPBRS. From left to right, the columns show the distributions of the height, uprightness, and joint-regularization rewards.}
    \label{rewarddistribution}
    \vspace{-15pt}
\end{figure*}
\begin{figure} 
    \centering
    \includegraphics[width=0.48\linewidth]{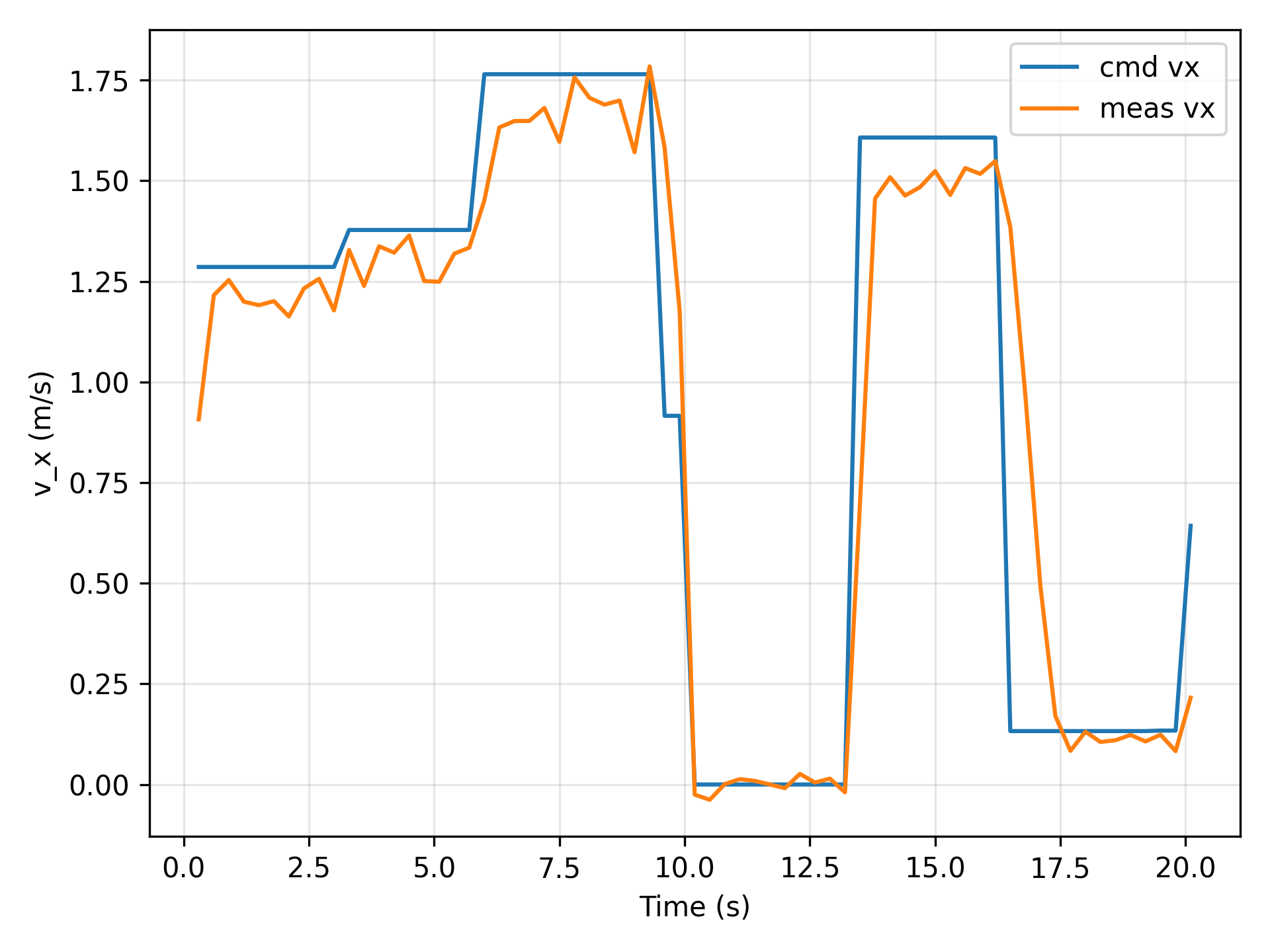}
    \includegraphics[width=0.48\linewidth]{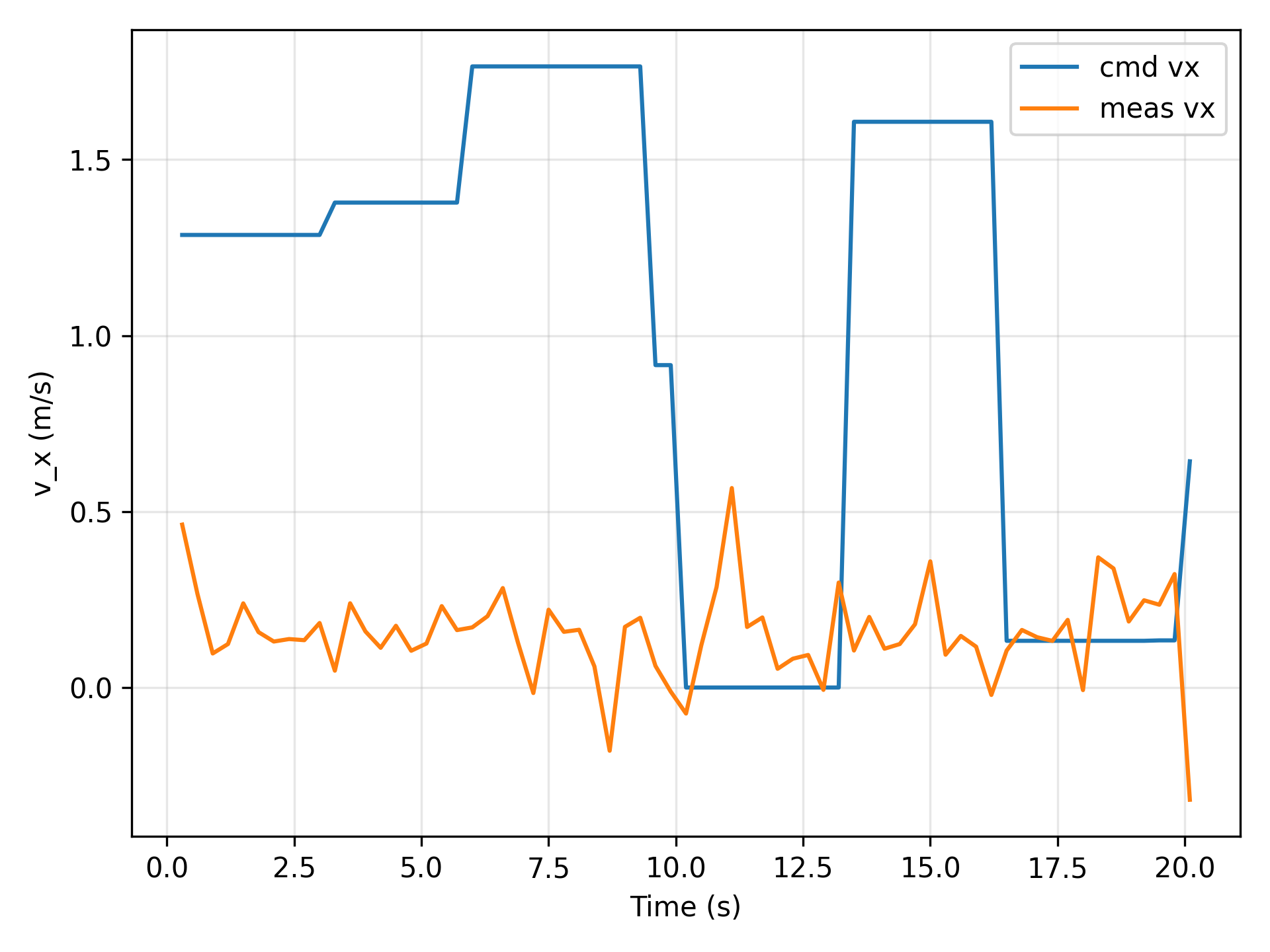}
    \includegraphics[width=0.48\linewidth]{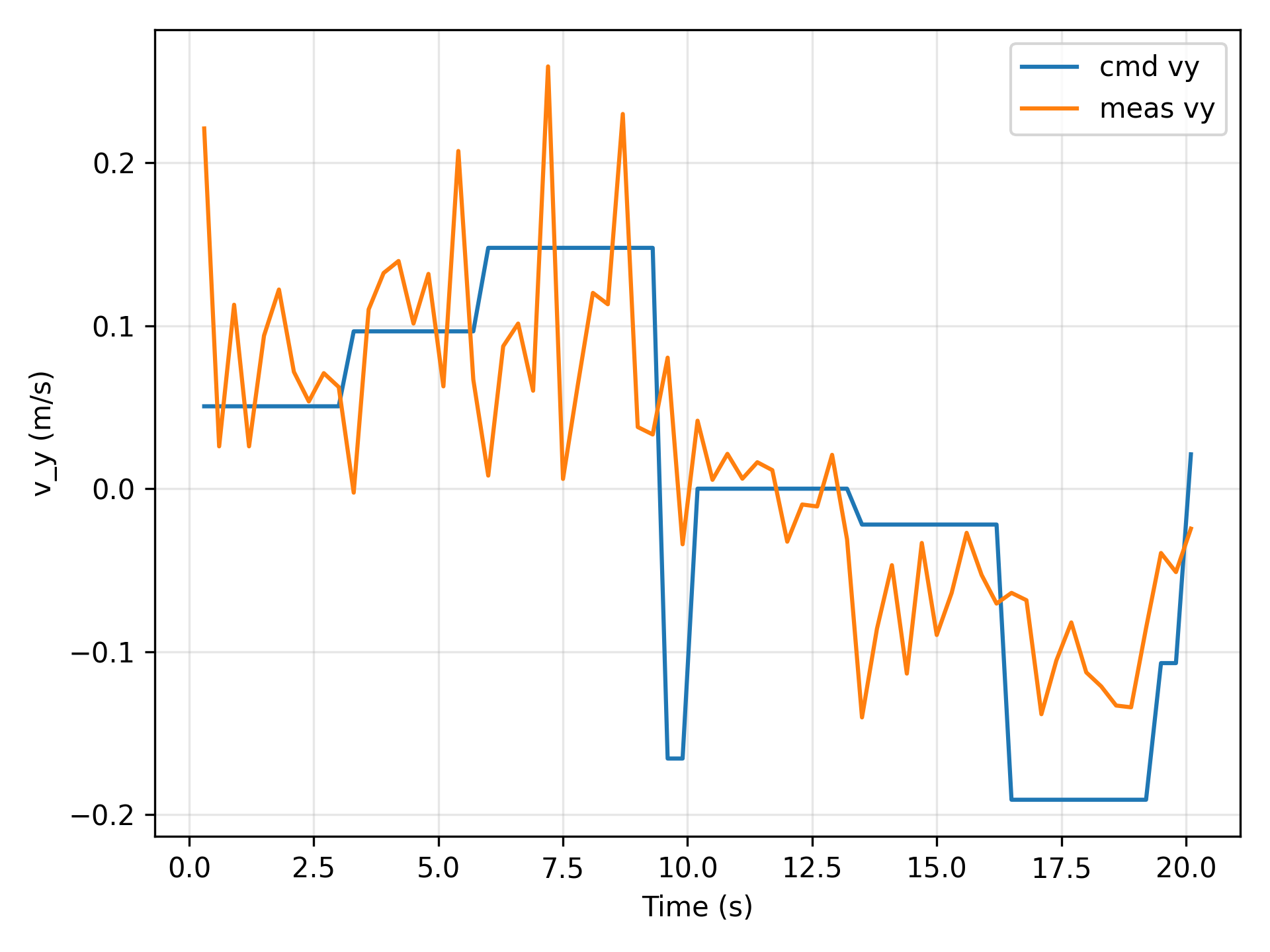}
    \includegraphics[width=0.48\linewidth]{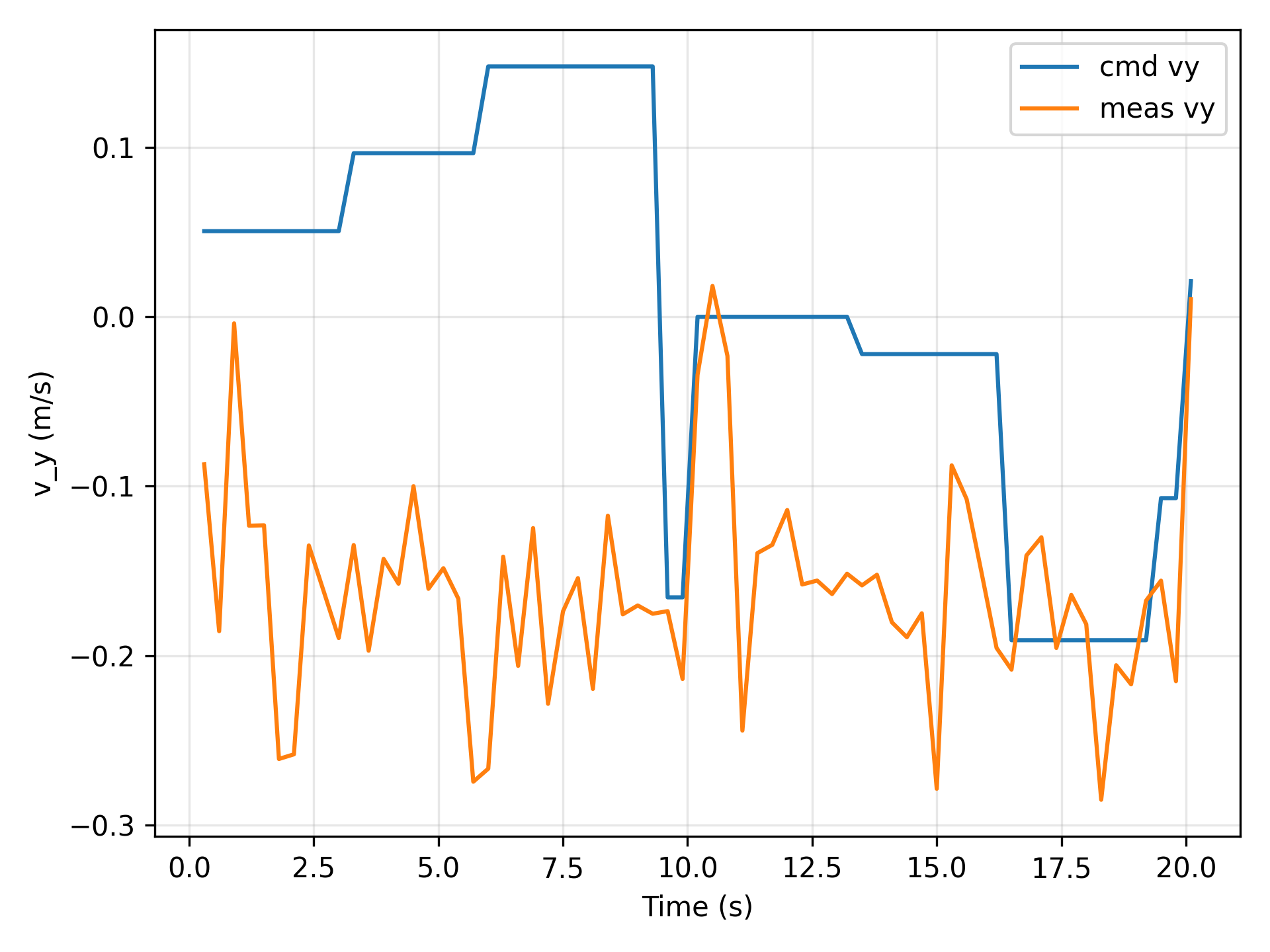}
    \includegraphics[width=0.48\linewidth]{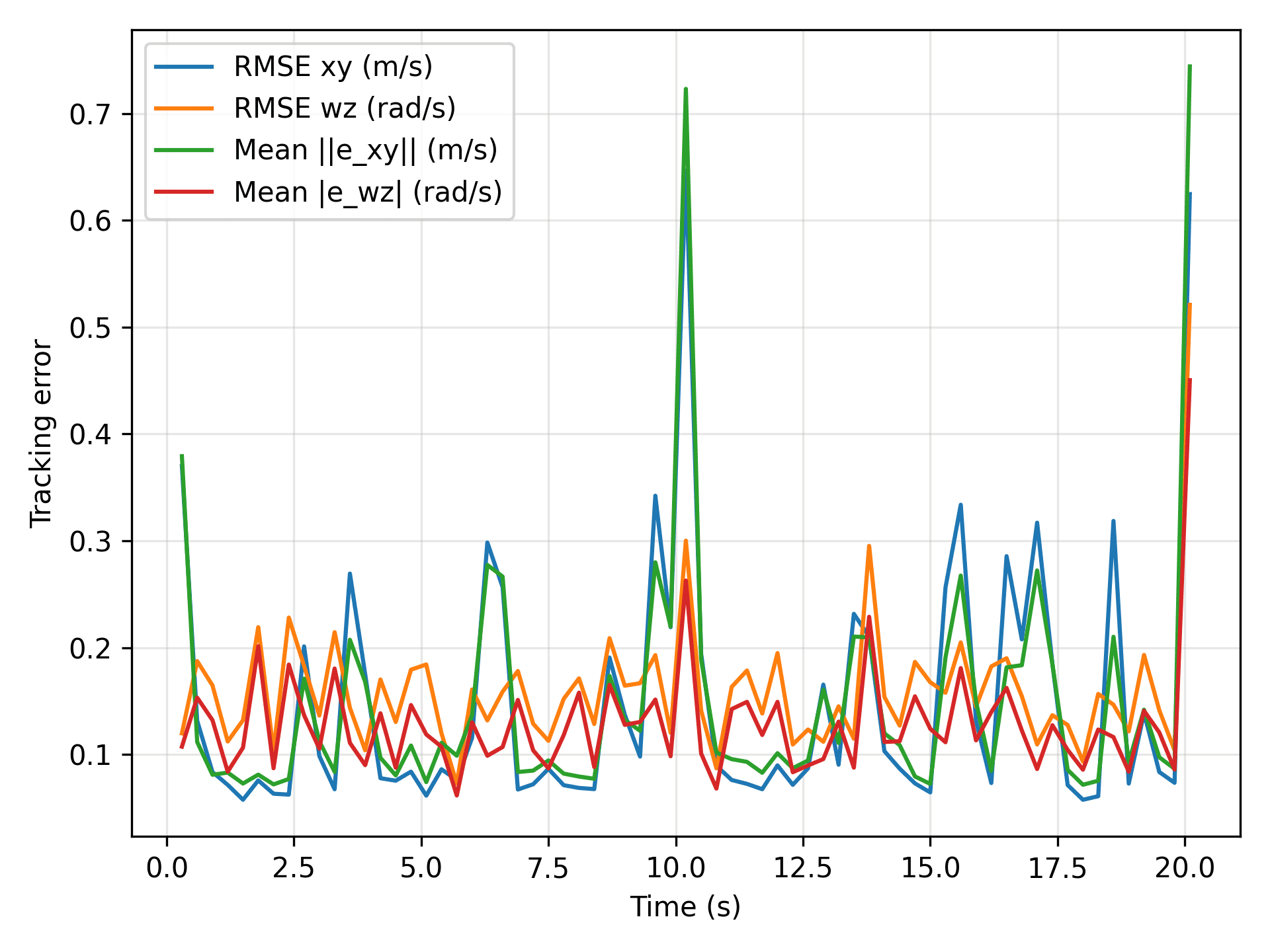}
    \includegraphics[width=0.48\linewidth]{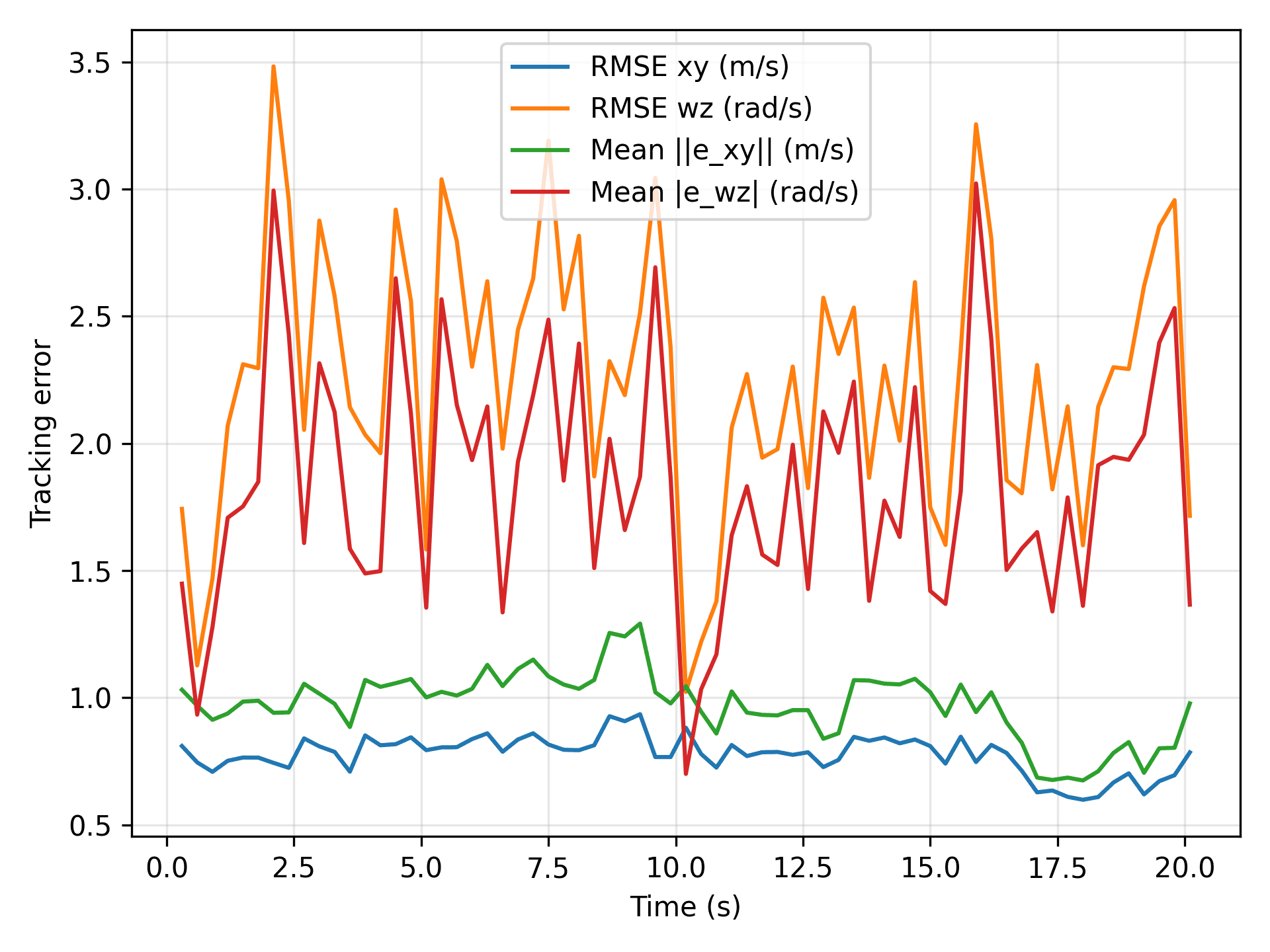}
    \caption{Tracking error of our method versus DRS with a shaping weight of 500. From top to bottom, the rows show tracking velocity in $x$, tracking velocity in $y$, and the mean error. While our method successfully tracks the commanded velocity, DRS with this weight causes the robot to stand still.}
    \label{trackingability500}
\end{figure}

In this section, we present results for the proposed method across nine humanoid robots using the same reward configuration, without any changes to the environment setup. We also benchmark our approach against two baselines: (i) the baseline locomotion reward and (ii) the baseline reward augmented with direct reward shaping (DRS) terms. For a fair comparison, we tune the baseline reward weights to achieve stable and reasonable locomotion performance, and we similarly tune the weights when adding the DRS terms. In contrast, for our method we keep the MPBRS weight fixed at $1.0$ for all robots, demonstrating that it does not require robot-specific reward tuning.
Finally, we compare both the cumulative reward and the reward command-tracking performance across all three methods to highlight the robustness of the proposed approach across different humanoid platforms. Robot trajectories are collected over $64$ timesteps. All agents are trained with randomized initial joint positions, where each joint is perturbed uniformly with an amplitude of $0.1$.
A detailed sensitivity analysis of the tuning weights was conducted on the Fourier N1 platform and is reported in the paper. For the remaining robots, we provide open-source code so that users can reproduce the experiments and benchmark performance across all humanoid platforms.
Open source code: \url{https://anonymous.4open.science/r/MPBRS-5448}.

\subsection{Training Benchmark}

We benchmark three training settings-Baseline, direct reward shaping (DRS), and MPBRS-across nine humanoid robots, as shown in Fig.~\ref{benchmarkcompare}. All three settings converge on all nine platforms. In the total-reward curves, DRS often appears to converge faster. This trend is expected because DRS adds auxiliary reward terms directly to the optimization objective, which increases the total reward signal. However, this also places DRS on a different reward scale than the Baseline and MPBRS settings. Therefore, total reward values are not directly comparable across all three settings, and DRS requires more careful reward-weight tuning.

Across training runs, MPBRS remains behaviorally consistent with the Baseline and does not produce noticeable changes in the learned policy. This is consistent with the PBRS formulation, where the shaping contribution approaches zero near the optimal state. In contrast, DRS introduces persistent auxiliary rewards into the objective. As a result, if the shaping weights are not carefully tuned, DRS can influence both the learned policy and the resulting robot behavior.

These effects are also reflected in the benchmark curves. MPBRS generally follows the Baseline trends, particularly in fall rate, whereas DRS shows larger deviations in total reward and, for some robots, in fall rate due to the added shaping terms. At the same time, all three settings achieve similar tracking performance overall, as indicated by comparable rewards for planar velocity tracking (\(x\)-\(y\)) and yaw-rate tracking (\(z\)-axis angular velocity). Thus, although DRS alters the reward landscape more strongly, the final tracking capability remains broadly comparable to the Baseline and MPBRS settings.

Another important observation is that convergence speed is strongly robot-dependent. Some humanoids improve rapidly and stabilize early, whereas others learn more gradually under all three settings. This indicates that morphology, actuator limits, and nominal robot dynamics remain dominant factors in training difficulty, even when the reward shaping strategy is changed. These results confirm that our method can run on multiple platforms without retuning the reward function and is also less sensitive to reward function tuning.

Overall, the benchmark highlights a practical trade-off. DRS can improve apparent learning speed, but it modifies the optimization objective and may introduce behavioral bias if the shaping weights are not tuned carefully. MPBRS provides a more conservative alternative: it improves training guidance while preserving baseline-like behavior and comparable tracking performance across a diverse set of humanoid robots.

We further evaluate the tracking performance of the policy trained with MPBRS on the Fourier N1 robot, as shown in Fig.~\ref{trackingabilityn1}. In this evaluation, the robot is respawned every 10 seconds to test whether the policy can recover and re-establish tracking after a reset. 

As shown in Fig.~\ref{trackingabilityn1}, the robot tracks the commanded planar velocities with good accuracy over the 17.5-second interval. In particular, the policy tracks the \(x\)-axis velocity command up to approximately \(3.5~\mathrm{m/s}\), while also following the \(y\)-axis velocity commands with reasonable accuracy despite larger fluctuations. The \(v_y\) response is noisier than \(v_x\), which is expected in humanoid locomotion due to lateral balance constraints; nevertheless, the measured velocity follows the command trend across multiple command changes.

A prominent feature in the figure is the tracking-error spike near the respawn time (around 10~s). These peaks are expected and are primarily caused by the reset event rather than by persistent degradation in policy performance. After each respawn, the tracking error decreases rapidly and returns to a low range, indicating that the policy can recover quickly and resume stable tracking. This behavior is also evident in the bottom panel: RMSE exhibits larger peaks because it is more sensitive to short transients, whereas the mean error remains comparatively small over most of the trajectory.

Overall, Fig.~\ref{trackingabilityn1} shows that MPBRS maintains strong velocity-tracking performance on the Fourier N1 robot. During normal operation, the tracking error stays low, while the transient spikes are predictable and mainly caused by the 10-second respawn protocol. The robot can track x-axis velocities up to 3.5~m/s with a small RMSE, which highlights the effectiveness of the method. The desired velocity is also changed during the test to demonstrate the policy’s adaptability.

More importantly, the response remains stable when the command changes, without large oscillations or long recovery periods. The error peaks that appear near respawn events are expected and should not be interpreted as instability of the policy. Instead, they come from the sudden resets in the evaluation setup. After each respawn, the robot quickly returns to the target velocity, showing that the policy can recover well from disturbances. These results suggest that MPBRS provides not only accurate tracking in nominal conditions, but also reliable and robust performance under repeated resets and changing commands.

\subsection{Sensitivity analysis}

In this section, we present a sensitivity analysis of the proposed method compared to the DRS. To analyze the sensitivity of the reward function tuning, we train the robot with different weight vectors for the reward. As shown in Figure~\ref{benchmarkweightsensitive} and Figure~\ref{rewarddistribution}, we vary the weights from 0.1 to 500 and benchmark the results to evaluate whether the robot successfully converges and tracks the command velocity. The total reward and tracking reward are presented in Figure~\ref{benchmarkweightsensitive}. The results show that even when the weight is set to 500, PBRS still helps the robot converge to the optimal policy and track the velocity command. In contrast, DRS is sensitive to the weight: when the weight is too high, the robot simply stands in one position, as the rewards for height, upright orientation, and joint regulation dominate other terms. The advantage of MPBRS is further illustrated in Figure~\ref{rewarddistribution}. When all rewards converge to zero, it indicates that the robot has reached the optimal state—meaning the rewards for height, upright posture, and joint regulation have achieved their optimal values. We further investigate the tracking ability of the robot in Figure~\ref{trackingability500} with a weight of 500. Our method successfully helps the robot find the optimal policy, whereas DRS only causes the robot to stand in one position.

\section{CONCLUSION}
This paper examined multi-potential-based reward shaping (MPBRS) for 3D humanoid locomotion using a single, unified reward configuration reused across nine humanoid robots. We combined standard direct shaping terms for command tracking and posture stabilization with three PBRS potentials: ray-based base-height regulation, gravity-alignment uprightness, and morphology-aware joint regularization (including ankle roll/pitch symmetry and hip-yaw penalties). Across diverse morphologies, we found that MPBRS provided only modest improvements in convergence speed. Its main benefit was practical: MPBRS was substantially less sensitive to reward scaling and weight choices, which simplified tuning and enabled more reliable transfer of the same reward design across different humanoids. Future work will investigate learned or adaptive potentials, improved sensor and noise modeling for sim-to-real robustness, and tighter integration with morphology-conditioned policies and domain randomization to further improve cross-robot generalization. 

\bibliographystyle{IEEEtran}
\balance
\bibliography{RefFile}
\end{document}